\documentclass{article}

\usepackage{iclr2026_conference,times}

\usepackage{amsmath,amsfonts,bm}

\def\eqref#1{equation~\ref{#1}}
\def\1{\bm{1}}

\DeclareMathAlphabet{\mathsfit}{\encodingdefault}{\sfdefault}{m}{sl}
\SetMathAlphabet{\mathsfit}{bold}{\encodingdefault}{\sfdefault}{bx}{n}

\usepackage[utf8]{inputenc}
\usepackage[T1]{fontenc}
\usepackage{hyperref}
\usepackage{url}
\usepackage{booktabs}
\usepackage{multirow}
\usepackage{amsfonts}
\usepackage{amsmath}
\usepackage{amssymb}
\usepackage{amsthm}
\usepackage{nicefrac}
\usepackage{microtype}
\usepackage{xcolor}
\definecolor{linkblue}{RGB}{0,102,204}
\usepackage{colortbl}
\usepackage{graphicx}
\usepackage{placeins}
\usepackage{capt-of}
\usepackage{wrapfig}

\title{%
  \hspace*{-0.em}%
  \raisebox{-0.018\textheight}{\includegraphics[width=0.06\textwidth]{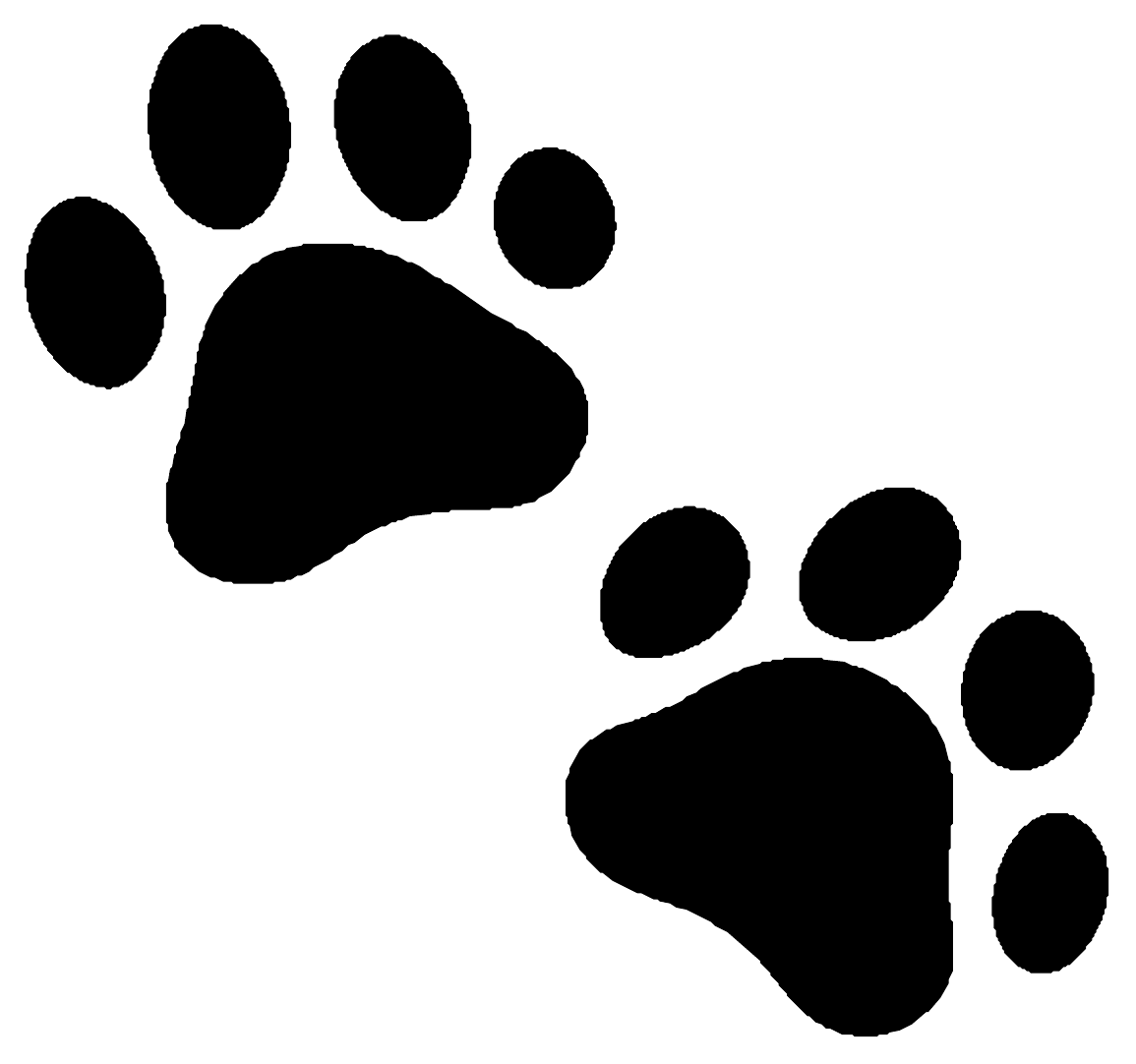}}%
  \hspace{0.2em}%
  PAWBench: How Far Are We from \\ Probabilistically Aligned World Modeling?%
}

\author{%
     \bf Yuandong Pu\textsuperscript{1,2} \thanks{This work was done during his internship at Shanghai Artificial Intelligence Laboratory.}
  ~~ \bf{Le Zhuo}\textsuperscript{3}
  ~~ \bf Sayak Paul\textsuperscript{4}
  ~~ \bf Gabriel Jorge Menezes \textsuperscript{3}
  \\ \bf Avram Đorđević \textsuperscript{3}
  ~~ \bf Shiyang Li \textsuperscript{2}
  ~~ \bf Yifan Zhou \textsuperscript{1}
  ~~ \bf Bin Fu \textsuperscript{2}
  ~~ \bf Wenlong Zhang \textsuperscript{2}
  \\ \bf Junjun He \textsuperscript{5}
  ~~ \bf Yu Qiao \textsuperscript{2}
  ~~ \bf Yihao Liu \textsuperscript{2} \thanks{Corresponding Authors}
  ~~ \bf Jinbo Xing\textsuperscript{6}
    ~~ \bf Xi Chen\textsuperscript7$^\dagger$
  \\
  \textsuperscript{1}Shanghai Jiao Tong University  ~~~
  \textsuperscript{2}Shanghai AI Laboratory  ~~~ 
  \textsuperscript{3} Krea AI~~~ \\
  \textsuperscript{4}Hugging Face ~~~
  \textsuperscript{5}Shanghai Innovation Institute ~~~
  \textsuperscript{6}Tongyi Lab ~~~
  \textsuperscript{7} The University of Hong Kong ~~~
}

\newcommand{\state}{x}
\newcommand{\action}{a}
\newcommand{\outcomes}{\mathcal{Y}}
\newcommand{\phat}{\hat{p}_{M}}

\iclrfinalcopy % Uncomment for camera-ready version, but NOT for submission.

\begin{document}

\maketitle

\begin{abstract}
Recent video generation models are increasingly framed as world models. Many physical processes can unfold in more than one valid way. Therefore, a world model should reproduce not only a plausible trajectory, but also the distribution of possible behaviors under the same initial observation and action. We call this distribution-level requirement \textbf{probabilistic alignment}.
However, existing evaluations largely assess individual-video plausibility and do not test whether repeated generations recover the correct distribution. This raises a central question: \textbf{how far are current video generators from probabilistically aligned world modeling?} To answer it, we formalize probabilistic alignment as a distributional criterion for world models and introduce \textbf{PAWBench}, a benchmark for evaluating video generators as stochastic samplers of world dynamics. We further introduce \textbf{PAWEval}, an outcome-level protocol that converts repeated video rollouts into empirical distributions over possible physical behaviors. Across 50 scenarios and eleven current systems, no model consistently matches the reference probabilities while recovering the range of valid behaviors. Having established this gap, we test whether language prompts, initial noise sampling, or model training can reshape the model's predictive distribution. We believe our work can serve as a foundation for future efforts to move towards probabilistically aligned world modeling.

\textbf{Project page:} \href{https://pawbench.github.io}{\textcolor{linkblue}{https://pawbench.github.io}}
\end{abstract}

\section{Introduction}
Recent advances in video generation have substantially improved instruction following, visual quality, and temporal coherence~\citep{google2025veo31fast,kling2026kling3,bytedance2026seedance25,lightricks2026ltx25,minimax2026h3}. Given an observation and a prompt or action-like control signal, current systems can generate plausible continuations of a scene. Together, these advances~\citep{openai2024sora,deepmind2025genie3,agarwal2026cosmos} have strengthened the case for viewing video generators not only as content creation tools, but also as visual world models.

Yet a world model must do more than render visually plausible rollouts. Because an observation reveals only part of the world state and an action does not specify all of its consequences, the generated rollouts must remain consistent with the scene's underlying geometry and dynamics \citep{li2026worldmodeltaxonomy}. When the underlying process is stochastic, the same initial observation and action can admit multiple physically valid futures rather than a single correct continuation \citep{babaeizadeh2018sv2p,denton2018svg}. A model should therefore capture the conditional distribution over these futures, not merely render one plausible sample. We call this requirement \textbf{probabilistic alignment}: under a fixed initial observation and action, the induced distribution should reflect both which futures are physically possible and how likely each one is. Probabilistic alignment matters for interaction and planning~\citep{ravi2025parallelworlds}, because decisions depend on both the possible consequences of an action and their relative likelihoods.

% Controlled processes such as dice, spinners, and collisions make this requirement measurable: their valid outcomes can be enumerated and, in calibrated cases, assigned analytically specified or symmetry-derived reference probabilities.

% Current video evaluation is still mostly organized around individual generations. Benchmarks measure visual quality, temporal coherence, text or action alignment, and physical plausibility \citep{vbench2024,vbenchpp2024,videophy2024,worldmodelbench2025}, all of which are necessary for world simulation. Yet success on these criteria does not reveal the model's underlying distribution over possible futures given the same observation and action. A model may still produce plausible individual videos while collapsing to a narrow subset of valid outcomes or assigning probability mass in the wrong proportions. As illustrated in Fig.~\ref{fig:one-plausible-future}, one plausible future is therefore not enough. This leaves a central question: \emph{how far are we from probabilistically aligned world modeling?}

\begin{figure}[t]
  \centering
  \includegraphics[width=\linewidth]{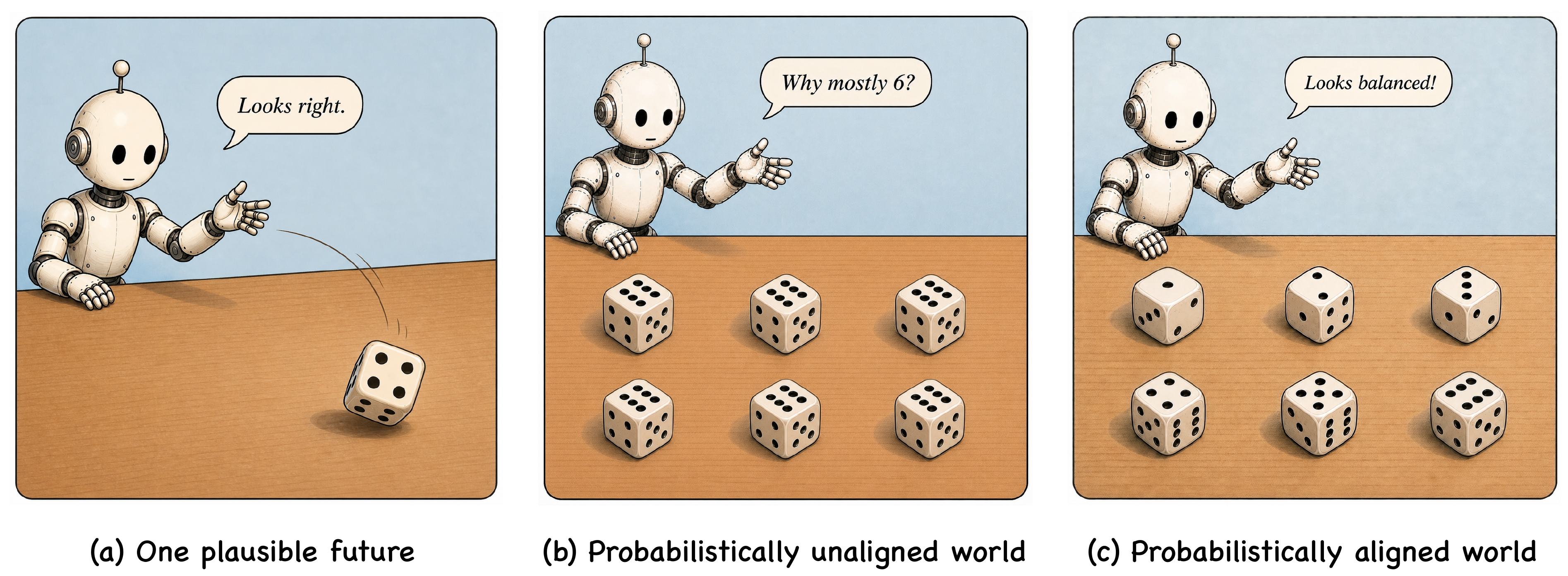}
  \caption{\textbf{One plausible future is not enough.} A single rollout from a video generation model can appear physically plausible, yet repeated rollouts from the same initial observation and action may reveal a probabilistically unaligned world: outcomes concentrate on a narrow subset of futures instead of matching the valid stochastic support. PAWBench evaluates this induced distribution or support over repeated futures, rather than judging a model by one sampled video.}
  \label{fig:one-plausible-future}
\end{figure}

Evaluation, however, still treats generated videos as isolated outputs rather than samples from the conditional distribution induced by the model. Existing benchmarks evaluate each generated video independently along dimensions of visual quality, temporal coherence, text or action alignment, and physical plausibility \citep{vbench2024,vbenchpp2024,videophy2024,worldmodelbench2025}. Yet success on these criteria does not reveal the model's underlying distribution over possible futures given the same initial observation and action. A model may still produce plausible individual videos while collapsing to a narrow subset of valid outcomes or assigning probability mass in the wrong proportions. As illustrated in Fig.~\ref{fig:one-plausible-future}, one plausible future is therefore not enough. This leaves a central question: \emph{how far are we from probabilistically aligned world modeling?}

To answer this question, we introduce \textbf{PAWBench} (Probabilistically Aligned World Bench), a diagnostic benchmark for evaluating whether video generators model the distribution of possible futures. PAWBench contains 50 scenarios spanning eight mechanism groups, organized into two complementary suites. \textit{PAW-Calibration} covers 25 scenarios with analytically specified or symmetry-derived reference distributions, such as tossing a coin or spinning a wheel. \textit{PAW-Coverage} covers 25 scenarios whose valid terminal outcomes can be enumerated but whose relative probabilities cannot be reliably specified, such as rolling a bowling ball or flipping a bottle. For each scenario, the source image and action prompt are held fixed across repeated rollouts. \textbf{PAWEval} maps readable, in-schema rollouts to terminal outcomes and records the remaining cases as outcome-readout failures. In PAW-Calibration, PAWBench compares the resulting empirical distribution with the reference probabilities; in PAW-Coverage, it measures recovery of the valid support.

% We benchmark current video generation models on PAWBench and find substantial gaps in both probability-mass alignment and support recovery. Tests with larger rollout budgets and human judgments indicate that these gaps are not solely artifacts of finite sampling or automated outcome readout. Further diagnostics show that models often fail to preserve valid physical trials, anticipate possible futures, or realize specified outcomes; their distributions also fail to track causal-state changes while shifting under non-causal cues. Together, these findings show that plausible video generation does not guarantee preservation of the conditional distribution over possible futures.
We benchmark eleven current video generation models on PAWBench. No model achieves all three requirements: accurate outcome probabilities, broad coverage of valid futures, and reliable performance across scenes. Tests with more rollouts support the sampling budget used in the main evaluation, while PAWEval agrees with human judgments on videos with clear terminal outcomes. Controlled interventions show that model distributions change too little when the physical transition changes, yet can shift when only a non-causal cue changes. Current video generators therefore remain far from probabilistically aligned world modeling: individual videos may look plausible even when their underlying distributions miss valid futures and assign them the wrong probabilities.

% This leaves open a more fundamental question: where should that distribution live? We ask whether stochasticity can be introduced through language, expressed through sampling noise, or learned by the model itself. At the language level, we ask whether external prompt control can reshape the distribution of futures produced by the generator. At the noise level, we ask whether stochasticity can be expressed more fully through sampling alone, coupling noise across rollouts to test whether the same budget spans a broader range of futures. At the model level, we intervene on the training distribution itself and ask whether the model internalizes not only which futures are possible, but how probability mass is distributed among them.
% Together, these probes distinguish stochasticity introduced at inference time from probability mass learned by the model itself.
We next move from measuring probabilistic alignment to asking whether it can be improved. We examine three points of intervention, moving from explicit language, through initial noise sampling, to the model’s learned predictive distribution under the same initial observation and action. Language makes possible futures explicit: VLMs predict what may happen, while prompts guide a video generator toward a named future. Sampling leaves the future unspecified and varies the initial noise to expose a broader range of possibilities from the same generator. Fine-tuning changes the model itself, testing whether it can learn how the probabilities of possible futures should vary with physical state.

Our main contributions are as follows.
\begin{itemize}

\item We formalize \emph{probabilistic alignment} as a distributional criterion
for world models and operationalize it through \textbf{PAWBench}, a 50-scenario
diagnostic benchmark spanning eight physical mechanism groups under fixed
initial observations and actions. \textbf{PAWEval} converts repeated video
rollouts into empirical outcome distributions, enabling PAW-Calibration and
PAW-Coverage to test probability-mass alignment and valid-support recovery,
respectively.

\item We benchmark current video generation models and reveal a failure hidden by current single-sample evaluation: plausible rollouts do not imply the correct distribution over possible futures. This gap persists beyond finite sampling and automated outcome readout.

\item We probe probabilistic alignment from external guidance to model
learning: we use language to predict and steer outcomes, sampling to broaden
finite-budget exploration, and fine-tuning to reshape the learned distribution.
These results distinguish inference-time steering and exploration from changes
to the model's learned predictive distribution.

\end{itemize}

\section{Probabilistically Aligned World Modeling}
\label{sec:capability}

\paragraph{World models represent distributions over possible futures.}
A world model should capture the range of futures that may unfold from a situation, not merely render one plausible continuation~\citep{li2026worldmodeltaxonomy}. Let $\state$ denote an initial observation and $\action$ an action. When the underlying process is stochastic, the same initial observation and action may lead to different physically valid outcomes. We therefore describe a video world model $M$ through the conditional distribution $P_M(\tau \mid \state,\action)$ over the future trajectories $\tau$ it generates. Probabilistic alignment concerns both which futures this distribution supports
and the relative probability it assigns to each one.

\paragraph{From a plausible future to an aligned distribution.}
Producing one plausible continuation meets only the weakest requirement: it shows that the model can realize one future. \textbf{\emph{Support alignment}} asks whether it can realize the distinct outcomes available under the same condition, rather than collapsing to a subset. \textbf{\emph{Probability-mass alignment}} is stronger: it asks whether those outcomes occur in the right proportions. A model can therefore be diverse without being aligned; covering every outcome does not guarantee assigning the correct probability to each.

% \paragraph{From trajectories to outcome distributions.}
To define these two levels of alignment, we map each generated trajectory to
the terminal outcome it realizes. Let $\outcomes$ be the finite set of terminal
outcomes, and let $g(\tau)\in\outcomes$ denote the outcome realized by
trajectory $\tau$. The model's trajectory distribution then induces
\[
p_M(y\mid\state,\action)
=
\Pr_{\tau\sim P_M(\cdot\mid\state,\action)}
\!\left[g(\tau)=y\right].
\]
This outcome distribution captures how much probability the model assigns to
each possible result, while abstracting away differences among trajectories
that reach the same result. When a reference distribution $q$ over $\outcomes$
is available, probability-mass alignment requires
$p_M(\cdot\mid\state,\action)=q$. When only the set of possible outcomes is
known, support alignment requires
$\operatorname{supp}\!\bigl(p_M(\cdot\mid\state,\action)\bigr)=\outcomes$.

% % Previous closing paragraph preserved verbatim for comparison; intentionally inactive.
% \iffalse
% \paragraph{What the criterion captures.}
% Outcome-level alignment separates distributional correctness from single-sample plausibility: a model may generate convincing individual rollouts while collapsing to one outcome or assigning incorrect relative probabilities. Support alignment rules out missing modes but not incorrect probabilities, making it necessary but insufficient whenever $q$ has support $\outcomes$. Outcome-level matching does not imply that the full trajectory distribution is correct. These are population properties; Section~\ref{sec:pawbench} describes how PAWBench estimates them from repeated rollouts.
% \fi

\section{PAWBench}
\label{sec:pawbench}

Having defined probabilistic alignment, we now describe how PAWBench makes this criterion measurable for video generators. The benchmark evaluates repeated rollouts under a fixed initial observation and action across two complementary alignment regimes, using an outcome-level evaluator. We first describe the overall benchmark design, including \textbf{PAW-Calibration} and \textbf{PAW-Coverage} (\S\ref{sec:pawbench:overview}), then present the scenario construction process that makes repeated-rollout statistics meaningful (\S\ref{sec:pawbench:scenarios}). We finally introduce \textbf{PAWEval}, a rubric-based outcome judging protocol that maps generated rollouts to terminal outcomes and converts them into distributional scores (\S\ref{sec:pawbench:paweval}).

\subsection{Benchmark Design}
\label{sec:pawbench:overview}
\begin{figure}[t]
  \centering
  \includegraphics[width=\linewidth]{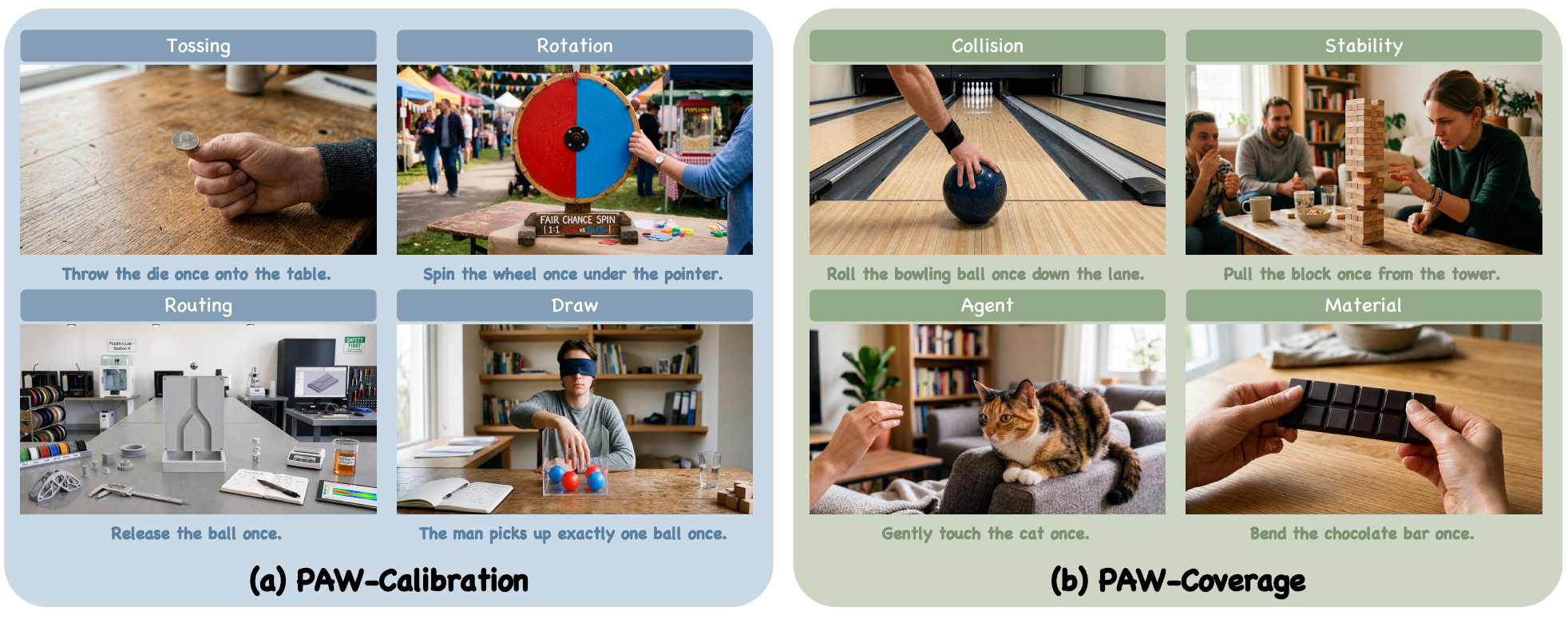}
  \caption{\textbf{PAWBench scenario taxonomy.} PAWBench covers eight mechanism groups under fixed initial observations and actions. PAW-Calibration contains calibrated probability scenarios with analytically specified reference distributions, while PAW-Coverage contains complex stochastic interactions evaluated by valid-support coverage. Each scenario fixes the source image and action prompt, then scores repeated model rollouts by their terminal outcomes.}
  \label{fig:scenario-taxonomy}
\end{figure}

PAWBench evaluates a video generation model through repeated rollouts conditioned on the same initial observation and action. Each benchmark item provides a source image representing the initial observation $\state$, an action prompt specifying the action $\action$, a finite set of valid terminal outcomes $\outcomes$, and, when available, a reference distribution $q$ over $\outcomes$. The model is queried $K$ times with the same $(\state,\action)$ pair, and the resulting videos are mapped to terminal outcomes and aggregated into an empirical outcome distribution $\phat$. Keeping the source image and action prompt fixed ensures that variation across rollouts reflects the model-induced future distribution rather than changes in the model inputs.

As shown in Fig.~\ref{fig:scenario-taxonomy}, PAWBench contains 50 scenarios spanning eight mechanism groups, divided into \textbf{PAW-Calibration} and \textbf{PAW-Coverage}. \textbf{PAW-Calibration} includes 25 scenarios whose valid terminal outcomes have analytically specified or symmetry-derived reference distributions, such as tossing, rotation, routing, and draw-style randomizers. It tests whether $\phat$ approaches the reference distribution $q$, or instead concentrates probability mass on a biased subset of valid futures. \textbf{PAW-Coverage} includes 25 scenarios whose valid outcomes can be enumerated but whose probabilities cannot be reliably specified, including collision, stability, agent interaction, and material transition scenarios. It tests whether repeated rollouts recover the qualitatively distinct valid futures in $\outcomes$. Detailed benchmark composition and outcome-set cardinalities are summarized in Fig.~\ref{fig:appendix-benchmark-statistics}.

This division separates two failures that single-sample evaluation cannot
distinguish. PAW-Calibration measures probability misallocation: every rollout may show a plausible die roll or spinner motion, yet one outcome may occur far too often. PAW-Coverage measures missing outcomes: individual rollouts may look plausible even though some valid futures never appear. We therefore report calibration and coverage separately rather than combine them into a single
score.

\subsection{Scenario Curation}
\label{sec:pawbench:scenarios}

PAWBench relies on repeated-rollout statistics, so each scenario must make the induced future distribution well defined. We therefore curate scenarios around three requirements. First, stochasticity must arise from a visible physical mechanism, rather than from ambiguity in the action prompt or hidden initial conditions. Second, the action prompt must specify one atomic intervention whose completion can be judged from the generated video. Third, the terminal outcomes must form a finite, visually distinguishable set so that rollouts can be consistently mapped into $\outcomes$. We manually curate and review each of the 50 scenarios against these requirements.

For each scenario, we generate one source image~\citep{google2025nanobananapro,openai2026gptimage2} to represent the initial observation and use one action prompt to specify the action across all rollouts. We list the valid terminal outcomes and specify when a rollout fails to complete the intended physical process. For PAW-Calibration, we also derive a reference distribution from analytic reasoning or physical symmetry. We review and finalize the source image, action prompt, outcome set, failure criteria, and reference distribution before evaluating any model. Appendix~\ref{app:scenario-construction} describes the construction process and the quality-control checks used to finalize each scenario.

\subsection{Evaluating Distributions over Possible Futures}
\label{sec:pawbench:paweval}

\begin{figure}[t]
  \centering
  \includegraphics[width=\linewidth]{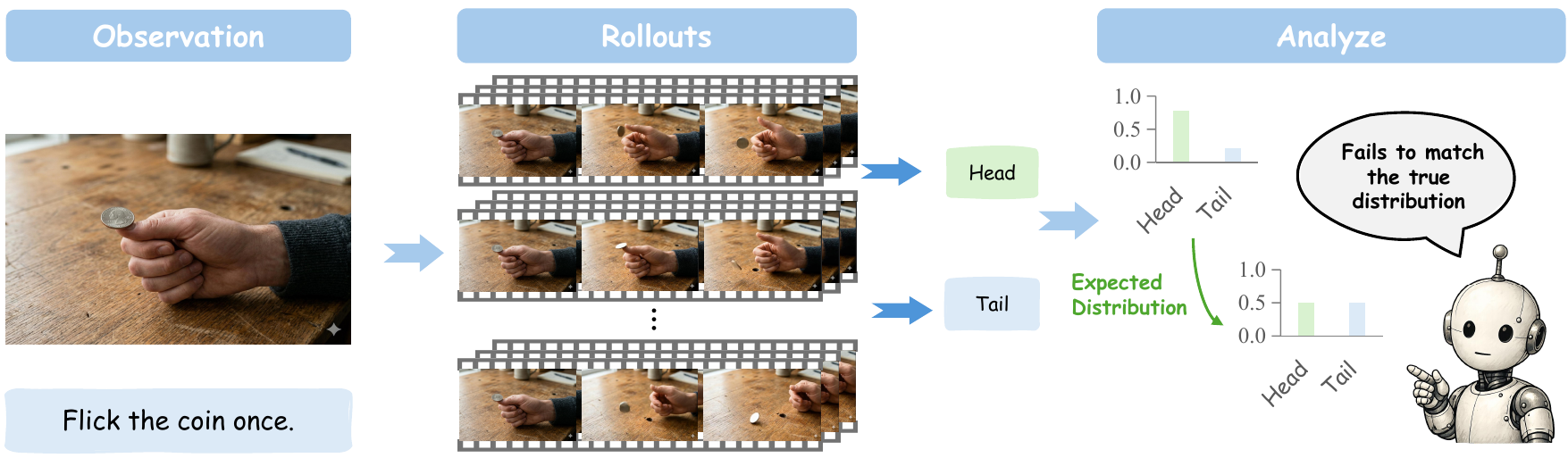}
  \caption{\textbf{PAWEval turns repeated rollouts into a distributional test.} In this PAW-Calibration example, a rubric-based judge maps each readable, in-schema coin-toss rollout to Head or Tail. Aggregating these labels yields the empirical outcome distribution, which is compared with the reference probabilities rather than matching generated videos frame by frame.}
  \label{fig:overview}
\end{figure}

To estimate the distribution induced by a video generator, we introduce
\textbf{PAWEval}, an outcome-level evaluation protocol for repeated rollouts. As illustrated in Fig.~\ref{fig:overview}, we generate $K$ video rollouts under the same $(\state,\action)$ pair. Each scenario has an outcome rubric that specifies which terminal physical result to identify, how valid results map to $\outcomes$, and when to return an outcome-readout failure. Gemini 3.5 Flash~\citep{google2026gemini35flash} applies this rubric to each rollout and returns either a terminal-outcome label in $\outcomes$ or an outcome-readout failure. We normalize the assigned outcome labels to obtain the conditional empirical distribution $\phat$; readout failures are reported separately because they do not represent physical outcomes in $\outcomes$ and cannot be counted as additional possible futures.

PAW-Calibration compares this conditional distribution with the reference
distribution $q$ using total variation distance (TVD), half the $\ell_1$
distance between the two categorical distributions~\citep{gibbs2002choosing}.
PAW-Coverage instead measures valid-support recovery as the fraction of valid
outcomes observed across repeated rollouts. Both metrics use readable,
in-schema outcomes; the outcome-readout gate separately determines whether a
scene is scoreable. Appendix~\ref{app:Evaluation} provides the full metric
definitions, gate, and aggregation rules used to compute the reported results.

\section{Evaluation on PAWBench}
\label{sec:experiments}

\subsection{Evaluation Setup}

We evaluate eleven current video generation models on both PAW-Calibration and PAW-Coverage, as listed in Tab.~\ref{tab:main}. The roster spans proprietary and openly released systems: HappyHorse~\citep{alibaba2026happyhorse}, Veo3.1 Fast~\citep{google2025veo31fast}, Kling 3 Std.~\citep{kling2026kling3}, Seedance 2~\citep{bytedance2026seedance2}, Wan2.7~\citep{alibaba2026wan27video}, Wan2.2~\citep{wan2025wan22}, LTX-2.3~\citep{ltx2026ltx23}, LTX-2.5~\citep{hacohen2026ltx2,lightricks2026ltx25}, Cosmos 3 Super I2V~\citep{nvidia2025cosmos,agarwal2026cosmos}, LingBot-Video-MoE~\citep{lingbot2026video}, and MiniMax H3~\citep{minimax2026h3}. For each PAWBench scenario, all systems receive the same source image and action prompt. In the main evaluation, we sample $K=50$ independent rollouts per scenario and system, with the action prompt held fixed across rollouts. Unless otherwise specified, we use each system with its default inference configuration. A scene passes the outcome-readout gate when no more than 30 of its 50 rollouts
fail outcome readout, equivalently when at least 20 yield readable, in-schema
outcomes. We report Scene Pass Rate (SPR) as the percentage of the 25 scenes in each track
that pass this gate.

% The separate trustworthiness audit records action completion, object continuity, and physical-process fidelity, but does not affect TVD, coverage, or Scene Pass Rate. Exact definitions are given in Appendix~\ref{app:metric-design}.

% \FloatBarrier
\subsection{Benchmark Results}
\label{sec:experiments:main}
\begin{table}[!htbp]
\centering
\scriptsize
\setlength{\tabcolsep}{0.7pt}
\renewcommand{\arraystretch}{1.06}
\definecolor{pawbest}{RGB}{244,214,210}
\definecolor{pawsecond}{RGB}{220,236,247}
\newcommand{\best}[1]{\cellcolor{pawbest}\textbf{#1}}
\newcommand{\second}[1]{\cellcolor{pawsecond}#1}
\newcolumntype{P}{>{\centering\arraybackslash}p{0.052\linewidth}}
\newcolumntype{R}{>{\centering\arraybackslash\normalfont\fontsize{6.4}{7.2}\selectfont}p{0.060\linewidth}}
\caption{\textbf{Main PAWBench results.}
PAW-Calibration reports conditional TVD ($\times100$, $\downarrow$): a 70/30
model distribution against a 50/50 reference scores 20, while an exact match
scores 0. PAW-Coverage reports valid-support recovery (\%, $\uparrow$). Avg. is computed over passing scenes;
SPR is the percentage of all 25 scenes that pass.
\textcolor{pawbest}{\rule{0.75em}{0.75em}} and
\textcolor{pawsecond}{\rule{0.75em}{0.75em}} indicate the best and second-best
conditional scores; / indicates that no scene in the corresponding mechanism
group passes the gate.}
\label{tab:main}
\vspace{0.35em}
\begin{tabular*}{\linewidth}{@{}p{0.195\linewidth}@{\extracolsep{\fill}}P R *{4}{P} P R *{4}{P}@{}}
\toprule
\multirow{2}{=}{\centering Model} & \multicolumn{6}{c}{PAW-Calibration: TVD $\times100$ ($\downarrow$)}
& \multicolumn{6}{c}{PAW-Coverage: Coverage (\%) ($\uparrow$)} \\
\cmidrule(lr){2-7}\cmidrule(lr){8-13}
& Avg. & SPR & Toss & Rot. & Rout. & Draw & Avg. & SPR & Coll. & Stab. & Agent & Mat. \\
\midrule
HappyHorse & 43.1 & 92.0\% & 40.1 & 27.5 & 41.9 & 59.5 & 47.1 & 100.0\% & 45.0 & 36.8 & 49.0 & 58.3 \\
Veo3.1 Fast & 35.4 & 88.0\% & 37.1 & 12.2 & 32.4 & 55.0 & 41.8 & 100.0\% & 39.8 & 21.1 & 53.3 & 44.2 \\
Kling 3 Std. & 34.9 & 92.0\% & 29.9 & 20.3 & 38.7 & 46.0 & 52.8 & 88.0\% & 52.7 & 40.0 & 44.3 & 77.5 \\
Seedance 2 & 30.5 & 100.0\% & 32.7 & 15.1 & 31.1 & 40.5 & 50.9 & 84.0\% & 48.6 & 50.0 & 44.3 & 67.5 \\
Wan2.7 & 26.3 & 92.0\% & 26.0 & 12.2 & 26.8 & 36.5 & 50.0 & 96.0\% & 60.3 & 50.5 & 37.9 & 53.3 \\
\midrule
Wan2.2 & 26.3 & 64.0\% & 22.4 & 15.8 & \second{23.2} & 45.9 & \second{63.4} & 92.0\% & \second{76.8} & \second{56.7} & 55.0 & 58.3 \\
LTX-2.3 & 30.1 & 24.0\% & 29.3 & \second{11.0} & / & 41.0 & \best{71.7} & 72.0\% & \best{82.7} & 50.0 & \best{59.5} & \best{90.0} \\
LTX-2.5 & 30.2 & 60.0\% & 30.2 & 17.1 & 39.9 & \second{27.1} & 57.4 & 88.0\% & 66.7 & 30.0 & \second{59.4} & 57.5 \\
Cosmos 3 Super I2V & \best{20.5} & 80.0\% & \best{19.4} & \best{3.6} & \best{17.7} & 35.5 & 55.2 & 92.0\% & 56.7 & 31.7 & 49.4 & \second{81.7} \\
LingBot-Video-MoE & 41.8 & 72.0\% & 31.8 & 13.4 & 44.4 & 60.8 & 58.8 & 100.0\% & 62.2 & \best{61.8} & 46.5 & 72.5 \\
MiniMax H3 & \second{24.2} & 68.0\% & \second{22.0} & 14.0 & 28.9 & \best{26.5} & 48.7 & 92.0\% & 59.5 & 39.5 & 36.4 & 58.3 \\
\bottomrule
\end{tabular*}
\vspace{-2.0em}

% \begin{flushleft}
% % \scriptsize
% % Avg. is computed over scenarios rather than mechanism columns. Rout., Coll., Stab., and Mat. denote routing, collision, stability, and material-transition scenarios.
% \end{flushleft}
\end{table}

\textbf{Current video generators remain far from probabilistically aligned world modeling.}
As shown in Tab.~\ref{tab:main}, no model combines well-aligned probability mass, broad support recovery, and reliable performance across scenes. Cosmos 3 Super I2V achieves the lowest Calibration TVD, but only 80.0\% of its Calibration scenes pass the outcome-readout gate. LTX-2.3 attains the highest Coverage average, yet that average is computed over the 72.0\% of its Coverage scenes that pass. Conversely, Seedance 2 and LingBot-Video-MoE pass every scene on Calibration and Coverage, respectively, but neither leads the corresponding conditional metric. Conditional alignment and scene-level reliability therefore capture distinct limitations of current models and must be read together.

\textbf{Recovering valid futures is not the same as assigning them the right probability.}
The models that recover broad support are not those that best match the reference probabilities, and this separation persists across physical mechanisms. A generator may expose many plausible outcomes while allocating probability mass incorrectly, or closely match the frequencies of observed outcomes while leaving valid alternatives unseen. PAW-Coverage and PAW-Calibration therefore expose complementary failures: recovering the valid support is necessary, but it does not establish that probability mass is correctly allocated among the outcomes within that support.

\begin{figure}[!htbp]
  \centering
  \includegraphics[width=\linewidth]{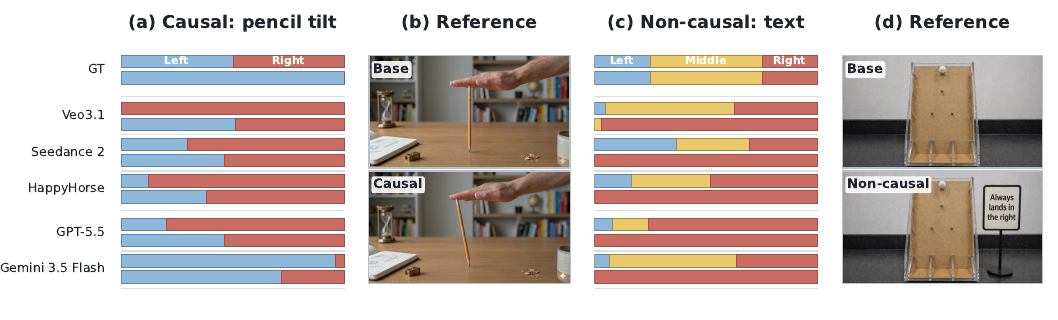}
\caption{\textbf{Models underreact to physically causal interventions and overreact to non-causal cues.} Upper and lower bars show outcome distributions before and after intervention; panels (b) and (d) show the paired scenes. The pencil tilt is causal because it changes the physical transition, whereas the Galton-board text is non-causal because it leaves the transition unchanged.}
  \label{fig:causal-cue-control}
\end{figure}

\textbf{Models Do Not Consistently Distinguish Causal from Non-Causal Changes.} A probabilistic world model should change its future distribution only when the physical transition changes. We test this criterion with paired interventions: physically causal interventions alter the transition and its reference distribution, whereas non-causal interventions change only an irrelevant visual or textual signal and should therefore leave both the physical process and its future distribution unchanged.

As shown in Fig.~\ref{fig:causal-cue-control}, model distributions shift incompletely or in the wrong direction under physically causal interventions. Under non-causal interventions, distractor text redirects probability mass despite an unchanged reference distribution. Appendix~\ref{app:causal-cue-full} reports the same pattern across all paired controls, video generators, and direct VLM future samplers. Models respond to altered inputs, but their responses do not reliably track whether the underlying physical transition has changed.

% \textbf{Strong conditional scores do not imply reliable alignment across scenes.}
% Tab.~\ref{tab:main} reports alignment scores averaged only over scenes that pass outcome readout; SPR instead records how often each model reaches that scoring regime across the full track. LTX-2.3, for example, achieves the highest conditional Coverage average but passes only 72.0\% of scenes; LingBot-Video-MoE recovers less support on average but passes all 25. Conditional score quality and benchmark-wide reliability therefore capture distinct dimensions of probabilistic alignment and must be interpreted together.

% \subsection{Diagnosing the Alignment Gap}

\subsection{Robustness of the Evaluation Protocol}
\label{sec:experiments:measurement}
\begin{figure}[t]
\centering
\setlength{\abovecaptionskip}{4pt}
\begin{minipage}[t]{0.52\linewidth}
  \centering
  \includegraphics[width=\linewidth]{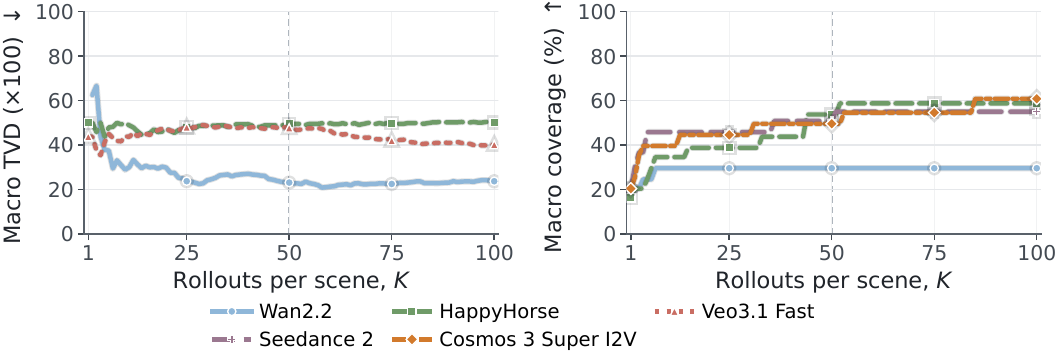}
  \captionof{figure}{
  \textbf{Larger rollout budgets increase coverage but leave calibration largely
  unchanged.} Conditional TVD (left) and valid-support coverage (right) as the
  rollout budget increases from $K=1$ to $100$.}
  \label{fig:sampling-budget-diagnostics}
\end{minipage}\hfill
\begin{minipage}[t]{0.46\linewidth}
  \centering
  \includegraphics[width=\linewidth]{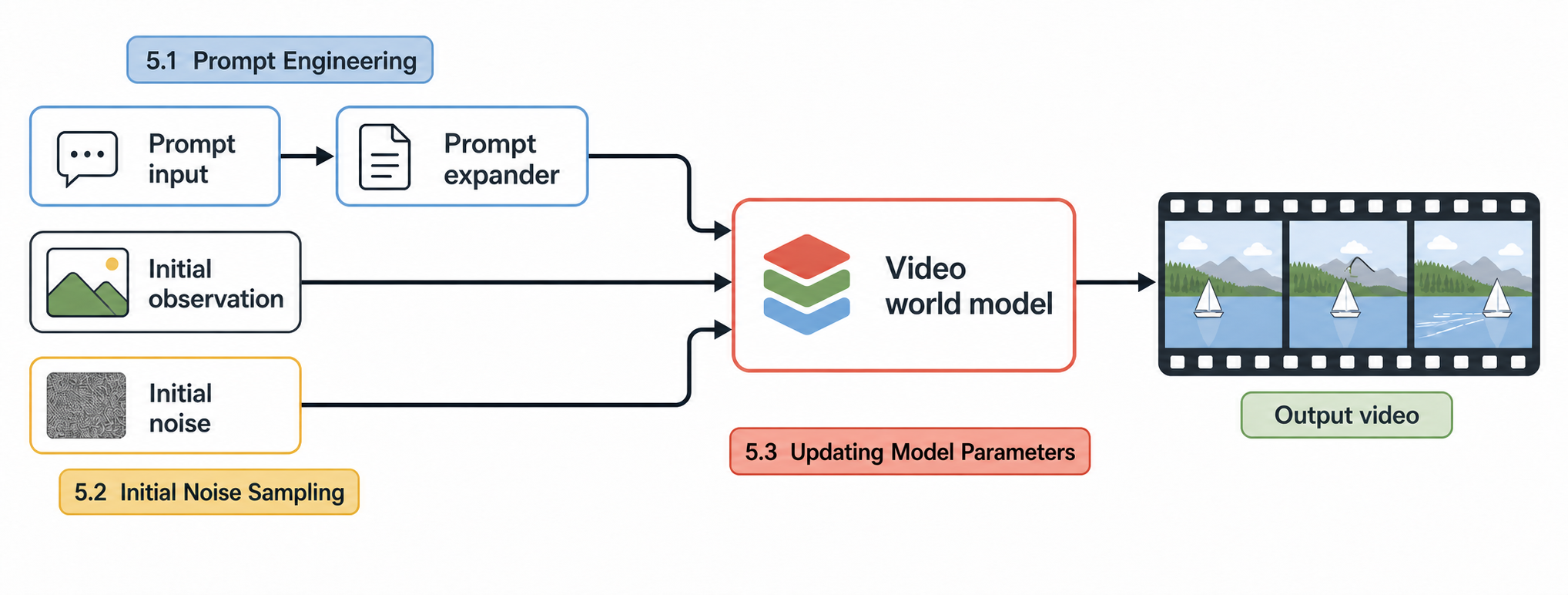}
  \captionof{figure}{
  \textbf{Three interfaces for shaping a world model's outcome distribution.}
  Section~\ref{sec:alignment-interventions} studies prompt engineering, initial
  noise sampling, and updating model parameters.}
  \label{fig:alignment-intervention-interfaces}
\end{minipage}
\end{figure}

The main results show that current models do not reproduce the relative
frequencies of valid futures and recover only part of the valid outcome
support. We test whether either finding reflects a limitation of the
evaluation protocol: too few rollouts or disagreement between PAWEval and
human judgments.
% \subsubsection{Does More Sampling Close the Gap?}

% \textbf{More sampling does not close the gap.} 
\textbf{Rollout-Budget Sensitivity.}
Figure~\ref{fig:sampling-budget-diagnostics} shows that doubling the rollout
budget leaves calibration largely unchanged, although coverage continues to
rise for three of the four models. Additional samples can uncover more valid
outcomes without correcting their observed relative frequencies. We therefore
use $K=50$ as a shared evaluation budget without treating it as a convergence
point. The Monte Carlo analysis in Appendix~\ref{app:finite-sample-null}
tests whether finite sampling could still account for the calibration gap.
Across the eleven video generators, observed TVD averages 31.2. For comparison,
we draw matched samples from the reference distributions while preserving each
model's passing scenes and readable sample counts. In 99\% of these simulations,
average TVD remains below 9.22, far below the calibration error observed across
the evaluated video generators.

% \subsubsection{Does PAWEval Agree with Human Judgments?}

% \textbf{Does PAWEval Agree with Human Judgments?} To distinguish model failures from readout failures, we ask whether PAWEval and humans assign the same terminal outcome to the same generated video. We collect seven independent human judgments per video using the same scene-specific outcome space. We compare the two on 888 videos for which both PAWEval and the human panel produce a clear terminal-outcome label. PAWEval matches the decisive human label on 722 videos (81.3\%). This agreement indicates that the PAWBench gap is not simply an artifact of PAWEval interpreting clear outcomes differently from humans. We provide the details in Appendix~\ref{app:human-study}.

\textbf{Agreement with Human Judgments.}
We collect seven independent human judgments for each video using the same
scene-specific outcome space. For the 888 videos where both PAWEval and the
human panel provide a clear terminal-outcome label, PAWEval agrees with the
decisive human label on 722 (81.3\%). Disagreement over clear outcomes therefore
cannot by itself account for the PAWBench gap.
Appendix~\ref{app:human-study} provides the full protocol and analysis.

% Having tested the two measurement explanations, we turn to the models themselves. Beyond the scored outcome distributions, the separate trustworthiness audit asks whether rollouts preserve the intended physical trial. We then test whether the outcome distribution responds to causal-state changes while remaining stable to irrelevant surface cues.

% \subsubsection{Do Rollouts Preserve the Intended Physical Process?}

% \subsubsection{Decomposing the World-Modeling Gap}

% The main PAWBench evaluation measures the full video-generation pipeline: given an initial state and an action, a model samples rollouts that induce a distribution over terminal outcomes. A mismatch in this distribution can arise from two necessary capabilities. The model may fail to infer which futures are possible and how likely they are, or it may infer a future but fail to realize that endpoint in video. We therefore separate the gap into \emph{semantic future prediction} and \emph{explicit endpoint realization}.

% For semantic future prediction, we remove video synthesis and ask multimodal models to predict directly in the canonical PAW outcome space using the same state-action inputs. This tests whether the action-conditioned future distribution is available before video decoding. For endpoint realization, we remove the need to choose among futures: video generators are given an explicit target outcome and evaluated by whether the generated terminal state matches the requested endpoint.

% \subsubsection{What Governs the Future Distribution?}

\section{Toward Probabilistically Aligned World Modeling}
\label{sec:alignment-interventions}

Generating one plausible future, or even controlling which future is produced, does not amount to controlling a distribution of futures. A probabilistically aligned world model must instead preserve the stochastic structure of possible outcomes under a fixed initial observation and action. For a video generator used as a world model, this distribution can be influenced at three levels: \textbf{prompt engineering}, \textbf{initial noise sampling}, and \textbf{updating model parameters}, as summarized in Fig.~\ref{fig:alignment-intervention-interfaces}. We therefore intervene at three interfaces that can shape this distribution: language specifies a requested future, initial noise sampling selects among futures, and model learning determines how probability mass is allocated among them. These probes distinguish inference-time steering and finite-budget exploration from changes in how the model itself distributes probability mass over future outcomes.

\subsection{Prompt Engineering}
\label{sec:alignment-language}

Language can steer a video generator by naming a future in its prompt. If a model can already render several plausible futures, varying these requests across rollouts could reshape its output distribution. We therefore evaluate both requirements: whether the language controller selects futures with the required frequencies or support, and whether the generator realizes each selected future.
\begin{center}
\centering
\begin{minipage}[t]{\linewidth}
  \vspace*{0pt}
  \centering
\captionof{table}{\textbf{VLM distributions over possible futures.}
Repeated VLM responses are mapped to PAWBench outcomes and aggregated into
empirical distributions without video generation.}
\label{tab:world-model-gap-diagnostics}

\scriptsize
\setlength{\tabcolsep}{0.7pt}
\renewcommand{\arraystretch}{1.08}
\definecolor{pawbest}{RGB}{244,214,210}
\definecolor{pawsecond}{RGB}{220,236,247}
\newcommand{\diagbest}[1]{\cellcolor{pawbest}\textbf{#1}}
\newcommand{\diagsecond}[1]{\cellcolor{pawsecond}#1}
\newcolumntype{Q}{>{\centering\arraybackslash}p{0.059\linewidth}}
\newcolumntype{S}{>{\centering\arraybackslash\normalfont\fontsize{6.4}{7.2}\selectfont}p{0.071\linewidth}}
\begin{tabular*}{\linewidth}{@{}p{0.205\linewidth}@{\extracolsep{\fill}}Q S *{4}{Q} Q S *{4}{Q}@{}}
\toprule
\multirow{2}{=}{\centering Model}
& \multicolumn{6}{c}{PAW-Calibration: TVD $\times 100$ ($\downarrow$)} & \multicolumn{6}{c}{PAW-Coverage: Coverage (\%) ($\uparrow$)} \\
\cmidrule(lr){2-7}\cmidrule(lr){8-13}
& Avg. & SPR & Toss & Rot. & Rout. & Draw & Avg. & SPR & Coll. & Stab. & Agent & Mat. \\
\midrule
Qwen3.5 Plus & 40.9 & 84.0\% & 36.4 & \diagbest{15.1} & 44.9 & 59.0 & 35.9 & 96.0\% & 19.8 & 49.5 & \diagsecond{48.6} & 29.2 \\
GPT-5.5 & 42.3 & 100.0\% & 35.6 & 42.9 & 41.9 & 56.0 & 34.3 & 100.0\% & 24.9 & 29.6 & 44.7 & \diagsecond{39.2} \\
GLM-5V Turbo & \diagbest{34.8} & 88.0\% & \diagbest{24.1} & 36.9 & \diagbest{33.9} & \diagbest{50.8} & 39.9 & 100.0\% & 30.9 & \diagbest{56.6} & 42.3 & 38.3 \\
Kimi K2.6 & \diagsecond{38.2} & 96.0\% & \diagsecond{32.6} & 32.1 & \diagsecond{36.8} & 58.5 & \diagsecond{45.1} & 100.0\% & \diagsecond{44.1} & \diagsecond{55.7} & 46.5 & 34.2 \\
Gemini 3.5 Flash & 39.4 & 100.0\% & 37.0 & \diagsecond{29.0} & 40.6 & \diagsecond{52.0} & \diagbest{46.6} & 96.0\% & \diagbest{49.5} & 39.5 & \diagbest{48.9} & \diagbest{43.3} \\
\bottomrule
\end{tabular*}

\end{minipage}
\end{center}

\begin{center}
\centering
\begin{minipage}[t]{\linewidth}
  \vspace*{0pt}
  \centering
\captionof{table}{\textbf{Predicted outcomes do not substitute for target outcomes.}
The first row scores the outcomes selected by GPT-5.5 PE before video synthesis.
Matched Base, PE, and Oracle PE results for four generators follow;
all conditions cover 25 scenes per PAWBench track at $K=50$.}
\label{tab:target-prompt-results}

\scriptsize
\setlength{\tabcolsep}{0.55pt}
\renewcommand{\arraystretch}{1.08}
\definecolor{targetbestshade}{RGB}{244,214,210}
\definecolor{targetsecondshade}{RGB}{220,236,247}
\newcommand{\targetBest}[1]{\cellcolor{targetbestshade}\textbf{#1}}
\newcommand{\targetSecond}[1]{\cellcolor{targetsecondshade}#1}
\newcolumntype{Q}{>{\centering\arraybackslash}p{0.059\linewidth}}
\newcolumntype{S}{>{\centering\arraybackslash\normalfont\fontsize{6.4}{7.2}\selectfont}p{0.071\linewidth}}
\begin{tabular*}{\linewidth}{@{}p{0.205\linewidth}@{\extracolsep{\fill}}Q S *{4}{Q} Q S *{4}{Q}@{}}
\toprule
\multirow{2}{=}{\centering Model}
& \multicolumn{6}{c}{PAW-Calibration: TVD $\times100$ ($\downarrow$)}
& \multicolumn{6}{c}{PAW-Coverage: Coverage (\%) ($\uparrow$)} \\
\cmidrule(lr){2-7}\cmidrule(lr){8-13}
& Avg. & SPR & Toss & Rot. & Rout. & Draw & Avg. & SPR & Coll. & Stab. & Agent & Mat. \\
\midrule
GPT-5.5 PE
  & 44.3 & 100.0\% & 40.1 & 49.5 & 43.6 & 49.0
  & 35.0 & 100.0\% & 35.4 & 34.6 & 35.5 & 33.3 \\
\midrule
Wan2.2
  & 26.3 & 64.0\% & 22.4 & 15.8 & 23.2 & 45.9
  & 63.4 & 92.0\% & 76.8 & 56.7 & 55.0 & 58.3 \\
\hspace{0.5em}$+$PE
  & 27.1 & 92.0\% & 23.5 & 15.4 & 28.8 & 39.5
  & 61.9 & 100.0\% & 68.1 & 50.7 & 62.9 & 57.5 \\
\hspace{0.5em}$+$Oracle PE
  & 15.3 & 84.0\% & \targetBest{10.5} & 13.9 & 14.0 & 27.3
  & 76.2 & 96.0\% & 71.1 & 72.7 & 82.8 & 76.7 \\
\midrule
Cosmos 3 Super I2V
  & 20.5 & 80.0\% & 19.4 & \targetBest{3.6} & 17.7 & 35.5
  & 55.2 & 92.0\% & 56.7 & 31.7 & 49.4 & \targetBest{81.7} \\
\hspace{0.5em}$+$PE
  & 31.6 & 92.0\% & 27.0 & 15.0 & 34.9 & 47.0
  & 64.9 & 100.0\% & 71.1 & 62.0 & 53.5 & 76.7 \\
\hspace{0.5em}$+$Oracle PE
  & \targetSecond{12.8} & 88.0\% & \targetSecond{17.6} & 9.2 & \targetSecond{13.2} & \targetBest{5.1}
  & \targetBest{87.3} & 100.0\% & \targetBest{84.8} & \targetBest{91.4} & \targetBest{91.3} & \targetSecond{80.8} \\
\midrule
MiniMax H3
  & 24.2 & 68.0\% & 22.0 & 14.0 & 28.9 & 26.5
  & 48.7 & 92.0\% & 59.5 & 39.5 & 36.4 & 58.3 \\
\hspace{0.5em}$+$PE
  & 38.9 & 96.0\% & 32.5 & 35.0 & 42.2 & 48.6
  & 49.8 & 96.0\% & 59.3 & 41.8 & 40.5 & 57.5 \\
\hspace{0.5em}$+$Oracle PE
  & \targetBest{10.8} & 96.0\% & 21.6 & \targetSecond{6.0} & \targetBest{2.1} & 12.6
  & \targetSecond{82.9} & 96.0\% & \targetSecond{82.4} & \targetBest{91.4} & 82.1 & 76.7 \\
\midrule
LTX-2.5
  & 30.2 & 60.0\% & 30.2 & 17.1 & 39.9 & 27.1
  & 57.4 & 88.0\% & 66.7 & 30.0 & 59.4 & 57.5 \\
\hspace{0.5em}$+$PE
  & 36.4 & 96.0\% & 34.3 & 28.7 & 36.6 & 48.0
  & 55.6 & 100.0\% & 59.8 & 42.1 & 51.4 & 68.3 \\
\hspace{0.5em}$+$Oracle PE
  & 18.0 & 88.0\% & 26.2 & 21.8 & 15.0 & \targetSecond{5.7}
  & 73.9 & 96.0\% & 56.0 & \targetSecond{78.0} & \targetSecond{89.9} & 73.3 \\
\bottomrule
\end{tabular*}

\end{minipage}
\end{center}

% \begin{table}[!t]
% \centering
% \input{layout/vlm_future_prediction_table}
% \input{layout/target_prompt_results_row}
% \end{table}

% We test three settings. First, we remove video generation and repeatedly ask VLMs~\citep{qwen2026qwen35,openai2026gpt55,zai2026glm5vturbo,moonshot2026kimi26,google2026gemini35flash} to predict what may happen from the same initial observation and action. We map each response to the PAWBench outcome space and aggregate the responses into an empirical distribution. Second, GPT-5.5 performs prompt engineering (PE) for each rollout: without access to the reference distribution, it selects a possible future from the initial observation and action and writes a generator prompt requesting that future. Third, for Oracle PE, we directly specify one target outcome in each generator prompt, arranging the targets across rollouts to follow the reference distribution in PAW-Calibration and to balance the valid outcomes in PAW-Coverage. Together, the three settings separate future selection from video realization. Direct VLM prediction tests selection alone, PE tests selection followed by generation, and Oracle PE supplies the target directly to test whether the generator can realize it.

We test three settings. First, prior work finds that language models often understand or describe a target distribution more accurately than they reproduce it through repeated sampling, although distribution-aware prompting and training can narrow this gap~\citep{meister2025distributional,gu2025dice,misaki2025ssot,sorensen2026spectrum}. We therefore ask whether VLMs can act as samplers over possible futures without being given the target probabilities. We repeatedly query five VLMs~\citep{qwen2026qwen35,openai2026gpt55,zai2026glm5vturbo,moonshot2026kimi26,google2026gemini35flash} under the same initial observation and action, map each prediction to the PAWBench outcome space, and aggregate the predictions into an empirical distribution. Second, GPT-5.5 performs prompt engineering (PE) for each rollout: without access to the reference distribution, it predicts a possible outcome and writes a generator prompt requesting that outcome. Third, for Oracle PE, we directly specify one target outcome in each generator prompt, arranging the targets across rollouts to follow the reference distribution in PAW-Calibration and to balance the valid outcomes in PAW-Coverage. Together, these settings separate two sources of error: predicting the wrong distribution of outcomes and failing to produce a requested outcome in video. Direct VLM future sampling tests the first, PE evaluates the full chain, and Oracle PE supplies the target outcomes directly to isolate the second.
\begin{wrapfigure}{r}{0.27\linewidth}
\vspace{-0.8em}
\centering
\includegraphics[width=0.96\linewidth]{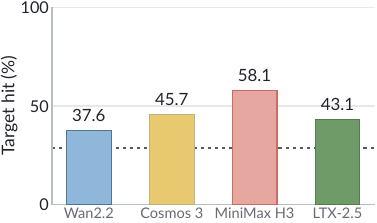}
\setlength{\abovecaptionskip}{2pt}
\caption{\textbf{Oracle PE often misses requested outcomes.}}
\label{fig:explicit-target-realization}
\vspace{-0.6em}
\end{wrapfigure}

The distributions produced by direct VLM sampling are already misaligned. Tab.~\ref{tab:world-model-gap-diagnostics} shows that GLM-5V Turbo achieves the lowest Calibration TVD at 34.8, while Gemini 3.5 Flash achieves the highest Coverage at 46.6\%. The outcomes selected by GPT-5.5 for PE are also misaligned before video generation, scoring 44.3 Calibration TVD and 35.0\% Coverage in Tab.~\ref{tab:target-prompt-results}. Passing these selections to video generators raises SPR for all four models, but increases Calibration TVD among passing scenes in every case and improves Coverage for only two. Manually specified targets perform better: Oracle PE lowers Calibration TVD and raises Coverage for every generator. Even with these targets, Fig.~\ref{fig:explicit-target-realization} shows that the generators realize only 37.6--58.1\% of the requested outcomes. Current language-based control is limited by both errors: the controller selects the wrong distribution of futures, and the generators often miss the supplied target. It can change individual rollouts without reliably aligning their distribution.

\subsection{Initial Noise Sampling}
\label{sec:alignment-noise}

The language interventions change which future is requested, but leave open what can be recovered when the request itself is held fixed. We therefore turn from language to sampling: noise determines which future is reached under the same request. Holding the action prompt and generator fixed, we test whether part of the observed gap is a finite-sample exploration failure, in which independent draws repeatedly visit the same modes even when other valid outcomes remain accessible.

We adopt Couple to Control (C2C)~\citep{jia2026coupletocontrol}, a repulsive Gaussian coupling scheme that introduces negative dependence among the \(K=50\) initial-noise samples while preserving each sample's standard Gaussian marginal. The action prompt, generator, and rollout budget remain fixed across C2C and independent sampling; Appendix~\ref{app:c2c-experimental-details} details the coupled-noise construction and the matched evaluation conditions used to isolate the effect of noise coupling.

\begin{center}
\centering
\begin{minipage}[t]{\linewidth}
  \vspace*{0pt}
  \centering
\captionof{table}{\textbf{Coupled noise broadens finite-gallery exploration.} C2C~\citep{jia2026coupletocontrol} couples $K=50$ rollout noises while
preserving the standard Gaussian marginal of each noise.}
\label{tab:c2c-sampling-results}

\scriptsize
\setlength{\tabcolsep}{0.55pt}
\renewcommand{\arraystretch}{1.08}
\definecolor{c2cbestshade}{RGB}{244,214,210}
\definecolor{c2csecondshade}{RGB}{220,236,247}
\newcommand{\cTwoCBest}[1]{\cellcolor{c2cbestshade}\textbf{#1}}
\newcommand{\cTwoCSecond}[1]{\cellcolor{c2csecondshade}#1}
\newcolumntype{Q}{>{\centering\arraybackslash}p{0.059\linewidth}}
\newcolumntype{S}{>{\centering\arraybackslash\normalfont\fontsize{6.4}{7.2}\selectfont}p{0.071\linewidth}}
\begin{tabular*}{\linewidth}{@{}p{0.205\linewidth}@{\extracolsep{\fill}}Q S *{4}{Q} Q S *{4}{Q}@{}}
\toprule
\multirow{2}{=}{\centering Model}
& \multicolumn{6}{c}{PAW-Calibration: TVD $\times100$ ($\downarrow$)}
& \multicolumn{6}{c}{PAW-Coverage: Coverage (\%) ($\uparrow$)} \\
\cmidrule(lr){2-7}\cmidrule(lr){8-13}
& Avg. & SPR & Toss & Rot. & Rout. & Draw & Avg. & SPR & Coll. & Stab. & Agent & Mat. \\
\midrule
Wan2.2
  & 26.3 & 64.0\% & 22.4 & 15.8 & 23.2 & 45.9
  & 63.4 & 92.0\% & 76.8 & 56.7 & 55.0 & 58.3 \\
\hspace{0.5em}$+$C2C
  & 25.7 & 84.0\% & \cTwoCBest{17.4} & 22.3 & 21.6 & 48.7
  & 69.2 & 88.0\% & 82.4 & \cTwoCSecond{57.5} & \cTwoCSecond{59.4} & 68.3 \\
\midrule
LTX-2.3
  & 30.1 & 24.0\% & 29.3 & 11.0 & / & 41.0
  & \cTwoCSecond{71.7} & 72.0\% & \cTwoCSecond{82.7} & 50.0 & \cTwoCBest{59.5} & \cTwoCBest{90.0} \\
\hspace{0.5em}$+$C2C
  & \cTwoCSecond{19.9} & 32.0\% & \cTwoCSecond{18.6} & 7.2 & 34.0 & \cTwoCSecond{27.7}
  & \cTwoCBest{74.8} & 72.0\% & \cTwoCBest{92.8} & \cTwoCBest{87.5} & 57.1 & 76.7 \\
\midrule
Cosmos 3 Super I2V
  & 20.5 & 80.0\% & 19.4 & \cTwoCBest{3.6} & \cTwoCSecond{17.7} & 35.5
  & 55.2 & 92.0\% & 56.7 & 31.7 & 49.4 & \cTwoCSecond{81.7} \\
\hspace{0.5em}$+$C2C
  & \cTwoCBest{19.4} & 72.0\% & 25.8 & \cTwoCSecond{6.1} & \cTwoCBest{14.3} & \cTwoCBest{24.2}
  & 63.9 & 92.0\% & 74.9 & 56.7 & 49.1 & 76.7 \\
\bottomrule
\end{tabular*}

\end{minipage}
\end{center}

Tab.~\ref{tab:c2c-sampling-results} reports results across all 25 scenes in each PAWBench track. Across the scenes that pass in each condition, C2C lowers mean Calibration TVD and raises mean Coverage for all three generators. The gains vary across mechanisms and do not consistently raise SPR. C2C therefore helps the 50 rollouts explore the model's existing possibilities more broadly, rather than changing the distribution learned by the model. We therefore turn next to the model itself.

\subsection{Updating model parameters}
\label{sec:alignment-model}

Unlike language and noise interventions, changing the training distribution can alter the generator's learned allocation of probability mass. We train five LoRA-adapted Wan2.2 models~\citep{hu2022lora} on training sets with different ratios of left- and right-falling pencil videos, with the left-fall share ranging from 0\% to 100\%, while holding the training budget and recipe fixed. Each model is evaluated on upright and left-leaning pencil scenes using a direction-neutral action prompt and $K=50$ rollouts. Because the interior mixtures contain both outcomes, differences among them probe relative mass rather than support. Appendix~\ref{app:training-intervention-details} provides the dataset, adaptation, and evaluation details.

\begin{center}
\centering
\newsavebox{\trainingtablebox}
\savebox{\trainingtablebox}{%
\begin{minipage}[t]{0.49\linewidth}
  \vspace*{0pt}
  \centering
\captionof{table}{\textbf{Training mixtures reshape outcome mass.}
TVD $\times100$ is reported for the unadapted Base and five LoRA models trained
on increasing proportions of left-fall examples.}
\label{tab:training-distribution-summary}

\small
\setlength{\tabcolsep}{3.8pt}
\renewcommand{\arraystretch}{1.42}
\definecolor{trainingbestshade}{RGB}{244,214,210}
\definecolor{trainingsecondshade}{RGB}{220,236,247}
\newcommand{\trainingBest}[1]{\begingroup\setlength{\fboxsep}{1.5pt}\colorbox{trainingbestshade}{\strut\textbf{#1}}\endgroup}
\newcommand{\trainingSecond}[1]{\begingroup\setlength{\fboxsep}{1.5pt}\colorbox{trainingsecondshade}{\strut#1}\endgroup}
\resizebox{0.98\linewidth}{!}{%
\begin{tabular}{@{}lccccccc@{}}
  \toprule
  \textbf{Scene}
    & \textbf{Reference}
    & \textbf{Base}
    & \multicolumn{5}{c}{\textbf{LoRA training $P(\mathrm{left})$ (\%)}} \\
  \cmidrule(lr){4-8}
  & \textbf{$L/R$} & & \textbf{0} & \textbf{20}
    & \textbf{50} & \textbf{80} & \textbf{100} \\
  \midrule
  Upright & 50/50 & 17.3 & 50.0 & \trainingBest{17.6} & \trainingSecond{23.5} & 48.0 & 50.0 \\
  Left-leaning & 100/0 & 41.3 & 97.2 & 50.0 & \trainingSecond{18.9} & \trainingBest{0.0} & \trainingBest{0.0} \\
  \bottomrule
\end{tabular}%
}

\end{minipage}%
}
\usebox{\trainingtablebox}\hfill
\begin{minipage}[t][\dimexpr\ht\trainingtablebox+\dp\trainingtablebox\relax][s]{0.49\linewidth}
  \vspace*{0pt}
  \vspace*{0.45em}
\centering
\includegraphics[width=0.84\linewidth]{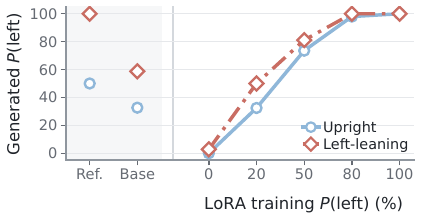}
\par\vfill
{\small
\setlength{\abovecaptionskip}{0pt}
\captionof{figure}{\textbf{Training steers outcome mass across scenes.}
Both scenes follow similar trends.}
\label{fig:training-mass-response}
}

\end{minipage}
\end{center}

Fig.~\ref{fig:training-mass-response} shows that increasing the share of
left-falling videos in the training set raises the generated left-fall
frequency in both scenes. The relationship is nonlinear: the generated
frequency does not follow the training proportion one-to-one. The consistent
shift shows that training-data composition can reshape the model's outcome
distribution; imbalanced outcome frequencies may therefore contribute to
current generators' distributional biases, although their actual training
distributions are unknown. A probabilistically aligned model should produce a
$50/50$ distribution for the upright pencil and a $100/0$ distribution for the
left-leaning pencil. None of the five adapted models produces both
(Tab.~\ref{tab:training-distribution-summary}). With 20\% left-falling videos
in training, the model comes closest to the upright reference, while the
left-leaning scene remains at $50/50$. Increasing that share to 80\% or 100\%
brings the left-leaning scene to its $100/0$ reference, but also makes the
upright pencil fall left almost every time. The same adjustment moves both
scenes in the same direction, so improving the match for one scene worsens the
other. Changing global outcome frequencies therefore provides only coarse
control over the model's distribution. Probabilistic alignment requires
learning how the distribution of possible futures should change with each
scene's initial physical state under a fixed action.

\section{Related Work}
\label{sec:related}

\paragraph{Video Generators as World Models.}
World models have long been used to support planning by predicting the consequences of actions, from classical model-based reinforcement learning to latent dynamics and decision-centric models \citep{sutton1990dyna,ha2018worldmodels,hafner2020dreamer,schrittwieser2020muzero}. Recent video systems increasingly pursue this role through controllable visual rollouts, including large-scale video generators and interactive world models \citep{openai2024sora,bruce2024genie,deepmind2025genie3,nvidia2025cosmos,yang2024unisim}. This shift makes repeated rollouts from the same initial observation and action natural. PAWBench focuses on the distributional question left open by these systems: not only whether a rollout is coherent, controllable, or visually realistic, but whether the induced samples place probability mass on the right futures.

\paragraph{Video-generation and world-model benchmarks.}
Video-generation benchmarks have substantially improved evaluation of visual quality, temporal consistency, and text-video alignment \citep{evalcrafter2023,vbench2024,vbenchpp2024}. Physics-focused benchmarks further probe physical plausibility and action following \citep{videophy2024,phygenbench2024,t2vphysbench2025,physvidbench2025, pu2026picabenchfarphysicallyrealistic,li2026acoustitraceplausiblesoundviolates}. Recent world-model benchmarks evaluate action-conditioned prediction, physical-law adherence, embodied consistency, or downstream utility for planning and control \citep{worldmodelbench2025,worldsimbench2024,physicalworldmodel2024,vp2_2023}. These evaluations are complementary to PAWBench, but their unit of analysis is usually an individual generation, a deterministic violation, or task success under a prompt. PAWBench instead holds the initial observation and action fixed, samples repeated rollouts, maps them to terminal outcome labels, and evaluates whether the empirical outcome distribution or support matches the stochastic structure of the scene.

\paragraph{Calibration and coverage under fixed actions.}
Calibration and distributional evaluation separate accuracy, fidelity, probability assignment, and support recovery \citep{guo2017calibration,nalisnick2019do,sajjadi2018pr,kynkaanniemi2019pr,naeem2020density}. Recent work on generative-video uncertainty also shows that confidence estimation is feasible and important \citep{mei2025squbed,mei2025c3}. Closest to our setting, CaliBench evaluates nine stochastic scenes with known reference distributions, reporting scorability separately from conditional TVD and testing for miscalibration~\citep{sadeghi2026calibench}. PAWBench shares the repeated-rollout, discrete-outcome perspective but separates two regimes: PAW-Calibration when defensible reference probabilities are available, and PAW-Coverage when only the valid support can be specified. It further pairs these measurements with controlled probes of causal state, language, sampling noise, and training distribution.
%Appendix~\ref{app:extended-related-work} provides a fuller discussion of related world-model lineages, video and physics benchmarks, uncertainty quantification, and distributional metrics.

\section{Limitations}
\label{sec:discussion}

PAWBench is designed as a diagnostic test of probabilistic alignment, and its current form has three main limitations. First, it evaluates stochastic futures through terminal outcomes, which makes distributional comparison tractable but does not fully capture trajectory-level dynamics or intermediate physical processes. Second, its estimates are based on a finite number of rollouts: larger sampling budgets can reveal the induced distribution more reliably, but they increase evaluation cost and do not by themselves correct biased model distributions. Third, PAWBench uses controlled, visually parseable scenarios to isolate stochastic future modeling, leaving longer-horizon, interactive, and embodied environments for future study. Future work should extend probabilistic alignment from terminal labels to richer state trajectories, study more efficient and reliable rollout-based estimators, scale the benchmark to interactive settings, and develop training objectives for models that explicitly learn calibrated distributions over possible futures across different physical states.

\section{Conclusion}
\label{sec:conclusion}

World modeling requires more than producing plausible continuations; a world
model should capture the distribution of possible futures under the same
initial observation and action. We introduce PAWBench to make this requirement
measurable through repeated video rollouts. Across eleven current video
generators, no model consistently matches the reference probabilities while
recovering the range of valid futures, and the gap cannot be explained by
finite sampling or disagreement with human judgments on clear outcomes.
Controlled interventions further show that model distributions do not reliably
track causal changes in the physical process. Language can request individual
futures, coupled noise can broaden finite-budget exploration, and fine-tuning
can shift outcome frequencies, but none reliably recovers the
scene-conditioned distribution over possible futures. PAWBench therefore shows
that plausible, diverse, or controllable rollouts do not by themselves
establish probabilistically aligned world modeling across different physical
states.

\clearpage
\bibliography{iclr2026_conference}
\bibliographystyle{iclr2026_conference}

\appendix
\clearpage
\section{Benchmark Details}
\label{app:benchmark-evaluation-details}

\subsection{Scenario Definition and Task Taxonomy}
\label{app:task-definition}

Each PAWBench scenario fixes the inputs and outcomes of one repeated-rollout
test, as shown in Tab.~\ref{tab:pawbench-task-taxonomy}. A source image
represents the initial observation, an action prompt specifies one atomic
intervention, and a finite outcome set $\outcomes$ lists the valid terminal
outcomes of the physical process. Each scenario also includes outcome-readout
criteria and, when it can be derived independently of model outputs, a
reference distribution $q$ over $\outcomes$. We finalize these elements before
evaluating any model.

\subsubsection{PAW-Calibration Tasks}

PAW-Calibration comprises scenarios whose reference distribution $q$ is fixed
before model evaluation. We derive $q$ from sector proportions, combinatorial
outcome counts, physical symmetry, or visible causal state---not from model
outputs or a default uniform assumption. When the initial observation and
action do not support a defensible reference distribution, the scenario is
assigned to PAW-Coverage or discarded. PAW-Calibration tests whether repeated
rollouts allocate probability mass in accordance with $q$.

\subsubsection{PAW-Coverage Tasks}

PAW-Coverage comprises scenarios whose valid terminal outcomes can be
enumerated but whose relative probabilities cannot be justified from the
initial observation and action. It tests support recovery across repeated
rollouts without assuming equal outcome probabilities.

\begin{table}[!t]
  \centering
  \small
  \setlength{\tabcolsep}{4pt}
  \renewcommand{\arraystretch}{1.14}
  \begin{tabular}{
    @{}
    >{\raggedright\arraybackslash}p{0.13\linewidth}
    >{\raggedright\arraybackslash}p{0.31\linewidth}
    >{\raggedright\arraybackslash}p{0.22\linewidth}
    >{\raggedright\arraybackslash}p{0.27\linewidth}
    @{}
  }
    \toprule
    Task family & Task definition & Terminal readout & Representative audit concerns \\
    \midrule
    \rowcolor{blue!8}
    \multicolumn{4}{@{}l}{\textbf{PAW-Calibration}} \\
    \textbf{Tossing}
    & An object is thrown, flicked, or released into one of several settled states.
    & Face, orientation, or resting region.
    & No throw or release; continuity break; settled state unreadable. \\
    \textbf{Rotation}
    & An object spins relative to a fixed partition, pointer, or orientation frame.
    & Final sector or orientation.
    & No rotation; reference frame or endpoint unreadable; out-of-schema orientation. \\
    \textbf{Routing}
    & A released object traverses a visible branching structure.
    & Terminal branch, lane, bin, pocket, or location.
    & No traversal; continuity break; terminal location unreadable or out of schema. \\
    \textbf{Draw}
    & One item is selected or released from a visible finite collection.
    & Selected identity or category.
    & No selection; item continuity break; selected identity unreadable or out of schema. \\
    \midrule
    \rowcolor{red!7}
    \multicolumn{4}{@{}l}{\textbf{PAW-Coverage}} \\
    \textbf{Collision}
    & An object undergoes contact or a target-directed interaction.
    & Contact, deflection, scoring, miss, or settling state.
    & Intervention absent; impossible transition; continuity or readability failure. \\
    \textbf{Stability}
    & A near-threshold arrangement is subjected to a perturbation.
    & Stable, shifted, toppled, or collapsed state.
    & No perturbation; continuity break; terminal state unreadable or out of schema. \\
    \textbf{Agent interaction}
    & A living agent receives a fixed visible stimulus.
    & Observable response category.
    & Stimulus not delivered as specified; agent continuity break; response unreadable. \\
    \textbf{Material transition}
    & A material or deformable object undergoes a fixed intervention.
    & Resulting physical or material state.
    & Intervention absent; material continuity break; final state unreadable or out of schema. \\
    \bottomrule
  \end{tabular}
  \caption{\textbf{PAWBench task taxonomy.} Each family identifies a recurring physical task, its terminal outcome readout, and representative outcome-readout or trustworthiness concerns. Outcome readout supplies the labels and scene gate used by PAWBench; trustworthiness checks are auxiliary diagnostics (Appendix~\ref{app:paweval-validity}).}
  \label{tab:pawbench-task-taxonomy}
\end{table}

% \subsubsection{Invalid Outcomes}

% PAWEval assigns invalid generations to a shared invalid outcome $\bot$ rather than treating them as additional stochastic modes. Invalid cases include unreadable endpoints, off-task clips, action failures, object-continuity breaks, physically impossible transitions, and out-of-schema terminal states. This mass remains part of the model-induced distribution, but it is excluded from valid diversity: in PAW-Calibration, $\bot$ receives zero reference probability; in PAW-Coverage, $\bot$ never counts as recovered support. Full metric definitions are given in Appendix~\ref{app:metric-design}.

\begin{figure}[!ht]
  \centering
  \includegraphics[width=0.95\linewidth]{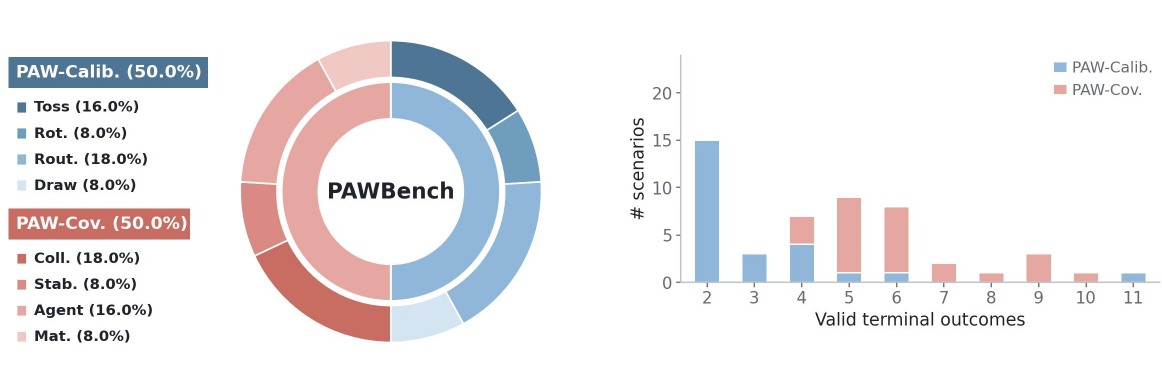}
  \caption{\textbf{Benchmark statistics of PAWBench.} PAWBench contains 50 manually curated scenarios, evenly divided between PAW-Calibration and PAW-Coverage, across eight stochastic mechanism groups. Each scenario fixes one source image and action prompt and uses a compact terminal-outcome set, allowing repeated rollouts to be aggregated into an empirical outcome distribution.}
  \label{fig:appendix-benchmark-statistics}
\end{figure}

\subsection{Scenario Construction and Quality Review}
\label{app:scenario-construction}

We construct each scenario so that variation across repeated rollouts reflects
the model's future distribution rather than ambiguity in the benchmark item.
Before evaluation, we finalize the source image, action prompt, valid terminal
outcomes, reference distribution when applicable, and outcome-readout criteria.
The physical mechanism must be visible, the action must specify one
intervention, and the terminal outcomes must be distinguishable.

\textbf{Source-image selection.} For each scenario, we generate candidate
source images with image generation models~\citep{google2025nanobananapro,openai2026gptimage2}
and manually select one image. The selected image must clearly show the
relevant physical mechanism and action target, avoid hidden initial conditions,
and support terminal outcomes that can be distinguished from the final frames.
We reject candidates with an obscured mechanism, an ambiguous initial state,
or competing action targets.

\textbf{Scenario finalization.} Given the selected source image, we write an
action prompt that specifies one atomic intervention whose completion can be
judged from the generated video. We then define the valid terminal outcomes,
criteria for readable in-schema labels, and separate trustworthiness checks.
For PAW-Calibration, we retain a reference distribution only when analytic
reasoning or physical symmetry justifies it independently of model outputs.
Outcome categories describe terminal physical states rather than incidental
differences in appearance or intermediate trajectories.

\textbf{Quality review.} We review each candidate as a complete scenario before
evaluating any model. The review checks mechanism visibility, action
specificity, terminal-outcome distinguishability, reference-distribution
justification when applicable, and the outcome-readout criteria. Failed checks
lead us to select or generate a new source image, revise the action prompt, or
refine the outcome categories. We discard a scenario when its ambiguity cannot
be removed without changing the physical process being evaluated.

\section{Evaluation Protocol and Supporting Analyses}
\label{app:Evaluation}
\subsection{PAWEval Outcome Readout}
\label{app:paweval-validity}

PAWEval uses Gemini 3.5 Flash~\citep{google2026gemini35flash} to apply a frozen
scene-specific outcome rubric to each generated rollout. The rubric returns a
terminal label $y\in\outcomes$ when the endpoint is readable and belongs to the
outcome set; otherwise, it returns the shared readout-failure label $\bot$.
When a visible endpoint is more specific than the outcome categories, the
rubric maps it to the corresponding category before aggregation. These outcome
labels determine conditional TVD and Coverage. Readout failures instead
determine whether a scene passes the outcome-readout gate and therefore affect
SPR; they never enter the conditional outcome distribution.

We also run an auxiliary trustworthiness audit that records whether the
requested action is completed, object continuity is preserved, and the
transition follows the specified physical mechanism. This separate diagnostic
does not change the terminal label or any PAWBench score.
Tab.~\ref{tab:paweval-validity-gates} summarizes the scoring readout and the
auxiliary audit.

For example, in a routing scene whose schema contains only named terminal bins, a ball that comes to rest visibly outside those bins has a readable but out-of-schema endpoint and is therefore recorded as an outcome-readout failure. By contrast, a spinner that ends clearly on blue after an implausible or discontinuous rotation still receives the in-schema label blue, while the trajectory is flagged only by the separate trustworthiness audit and the blue label still enters the conditional outcome counts.

\begin{table}[!htbp]
  \centering
  \small
  \setlength{\tabcolsep}{5pt}
  \renewcommand{\arraystretch}{1.14}
  \begin{tabular}{
    @{}
    >{\raggedright\arraybackslash}p{0.19\linewidth}
    >{\raggedright\arraybackslash}p{0.37\linewidth}
    >{\raggedright\arraybackslash}p{0.37\linewidth}
    @{}
  }
    \toprule
    Criterion & Required condition & Failure condition \\
    \midrule
    \rowcolor{blue!8}
    \multicolumn{3}{@{}l}{\textbf{Outcome readout}} \\
    \textbf{Endpoint readability}
    & The terminal state can be determined visually.
    & The endpoint is hidden, ambiguous, or unreadable. \\
    \textbf{Schema membership}
    & The terminal state maps to the fixed outcome set $\outcomes$.
    & A readable endpoint falls outside $\outcomes$. \\
    \midrule
    \rowcolor{red!7}
    \multicolumn{3}{@{}l}{\textbf{Trustworthiness audit} \normalfont} \\
    \textbf{Action completion}
    & The requested intervention is executed as specified.
    & The action is absent or the clip is off task. \\
    \textbf{Object continuity}
    & Key objects remain identifiable throughout the rollout.
    & Objects disappear, duplicate, or change identity. \\
    \textbf{Physical process}
    & The transition follows the specified physical mechanism.
    & The transition violates the specified physical mechanism. \\
    \bottomrule
  \end{tabular}
  \caption{\textbf{PAWEval separates scoring from auxiliary diagnostics.} Outcome readout supplies the labels and scene gate used by PAWBench; trustworthiness criteria only diagnose how readable outcomes are realized.}
  \label{tab:paweval-validity-gates}
\end{table}

\begin{figure}[!t]
  \centering
  \includegraphics[width=1\linewidth]{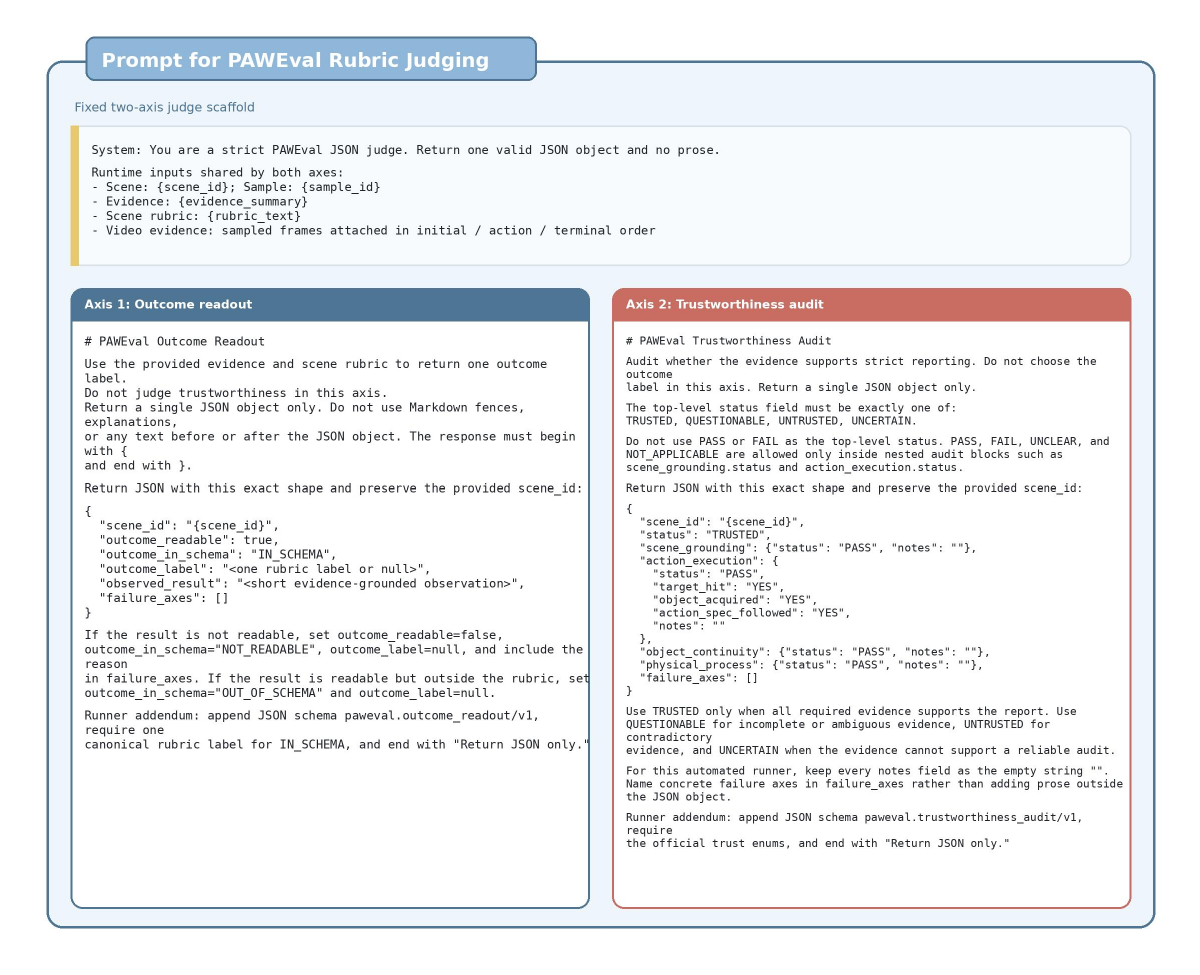}
  \caption{\textbf{PAWEval prompt scaffold.} The scene-specific outcome rubric assigns a terminal label or an outcome-readout failure. A separate trustworthiness rubric checks action execution, physical process, and object continuity without changing the PAWBench score. The placeholders \texttt{\{evidence\_summary\}} and \texttt{\{rubric\_text\}} are filled with sampled-frame evidence and the scene-specific rubric.}
  \label{fig:paweval-rubric-prompt}
\end{figure}

Outcome-readout failures remain separate from the conditional distribution
over readable, in-schema outcomes. Fig.~\ref{fig:paweval-rubric-prompt} shows
the fixed PAWEval prompt scaffold into which the scene-specific outcome and
trustworthiness rubrics are inserted.

\subsection{Auxiliary Trustworthiness Diagnostics}

The trustworthiness audit asks how a readable endpoint was reached. A clip may
reach such an endpoint without completing the intended physical process: the
action may not be executed, object identity may break, or the endpoint may be
reached through physically implausible dynamics. These flags do not change the
outcome label, TVD, Coverage, or SPR.

\begin{figure}[!t]
  \centering
  \includegraphics[width=0.60\linewidth]{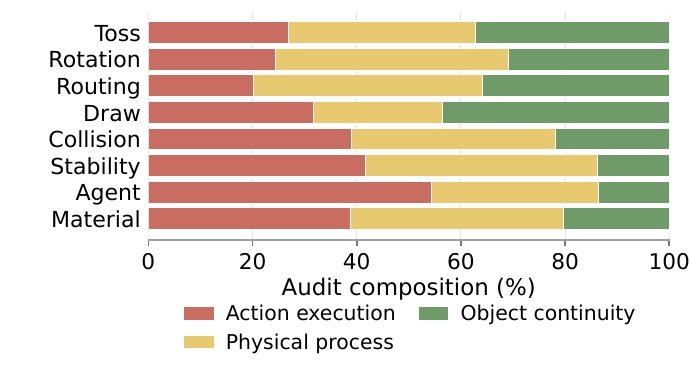}
  \caption{\textbf{Failure anatomy.} Incidence composition by mechanism group for eight models: HappyHorse, Veo~3.1 Fast, Kling~3 Std., Seedance~2, Wan2.7, Wan2.2, LTX-2.3, and Cosmos~3 Super I2V. Percentages use classifiable audits and count non-exclusive failures.}
  \label{fig:failure-anatomy}
\end{figure}

The audit covers three types of failure: action execution, physical process,
and object continuity. Fig.~\ref{fig:failure-anatomy} shows that
physical-process errors account for the largest share overall, while
action-execution and object-continuity errors remain substantial across
mechanism groups. The figure therefore describes how readable endpoints are
reached; it does not change the scores in Tab.~\ref{tab:main}.

\subsection{Metric Computation and Aggregation}
\label{app:metric-design}

PAWBench first applies a scene-level outcome-readout gate. For each model--scene pair, we evaluate $K=50$ rollouts generated under the same initial observation and action. The outcome readout assigns each rollout either a terminal label $y_i\in\outcomes$ or the readout-failure label $\bot$. Let $m$ denote the number of rollouts assigned to $\bot$. A scene passes the gate when $m\leq30$, equivalently when at least 20 rollouts have readable, in-schema outcomes. Otherwise, the scene does not contribute a conditional alignment score.

Conditional metrics are computed only for passing scenes. Let $n_{\mathrm{readout}}=K-m$. The empirical distribution over readable, in-schema outcomes is
\[
\hat{p}_{M,\mathrm{readout}}(y)
=\frac{1}{n_{\mathrm{readout}}}
\sum_{i=1}^{K}\mathbb{1}[y_i=y],
\qquad y\in\outcomes.
\]
The readout-failure label $\bot$ is not included in this conditional distribution.

For PAW-Calibration, each scenario specifies a reference distribution $q$ over its valid terminal outcomes. We measure valid-only conditional total variation distance,
\[
\mathrm{TVD}_{\mathrm{readout}}
(\hat{p}_{M,\mathrm{readout}},q)
=\frac{1}{2}
\sum_{y\in\outcomes}
\left|\hat{p}_{M,\mathrm{readout}}(y)-q(y)\right|.
\]
Lower TVD indicates closer probability-mass alignment among readable, in-schema outcomes. We report $100\times\mathrm{TVD}$.

For PAW-Coverage, the valid outcome set is enumerable but its probabilities are not specified. We measure valid-support recovery,
\[
\mathrm{Cov}_{\mathrm{readout}}
(\hat{p}_{M,\mathrm{readout}},\outcomes)
=\frac{
\left|\{y\in\outcomes:\hat{p}_{M,\mathrm{readout}}(y)>0\}\right|
}{|\outcomes|}.
\]
Higher coverage indicates that repeated rollouts recover a broader set of valid terminal outcomes. We report coverage as a percentage. Because observed support depends on the rollout budget, all primary comparisons fix $K=50$; Fig.~\ref{fig:sampling-budget-diagnostics} examines sensitivity to increasing the budget to $K=100$.

Mechanism-group scores average the scene-level conditional metric over passing scenes in that group. The track-level \emph{Avg.} instead averages over all passing scenes in the corresponding 25-scene track; it is not the simple average of the four mechanism-group scores. Scene Pass Rate is $n_{\mathrm{pass}}/25$, where the denominator includes the complete track roster. Conditional alignment scores and Scene Pass Rate should therefore be read together. PAW-Calibration and PAW-Coverage remain separate and are not combined into a single ranking.

\FloatBarrier
\subsection{Finite-Sample Effects on Calibration}
\label{app:finite-sample-null}

Because PAW-Calibration estimates an outcome distribution from a finite number
of readable rollouts, even samples drawn from the reference distribution $q$
can produce nonzero TVD. We measure this sampling error with matched Monte
Carlo simulations. Models pass different sets of scenes and yield different
numbers of readable rollouts, so we construct a separate matched
baseline for each model. For each passing model--scene pair, we draw the same
number of readable outcomes from the scene's reference distribution and compute
TVD using the same aggregation as in the main evaluation. These simulations
preserve each model's passing scenes and readable sample counts, replacing only
the generated outcomes with samples from $q$.

\begin{figure*}[!t]
  \centering
  \includegraphics[width=0.94\textwidth]{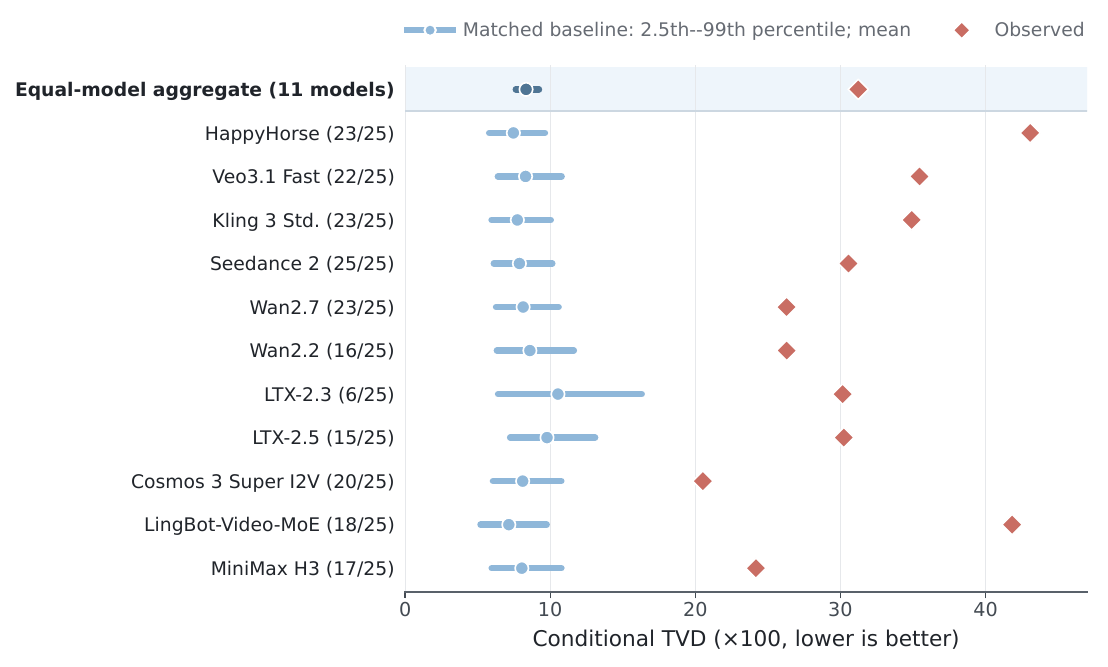}
  \caption{\textbf{Observed TVD exceeds the matched finite-sample baseline.}
  Blue intervals show the 2.5th to 99th percentiles obtained by sampling from
  the reference distributions with each model's passing scenes and readable
  sample counts. Circles mark the simulated means, and red diamonds mark the
  observed model TVDs. The aggregate row first averages passing scenes within
  each model and then weights the eleven models equally.}
  \label{fig:finite-sample-null}
\end{figure*}

Figure~\ref{fig:finite-sample-null} compares the observed model TVDs with these
matched baselines. Across the eleven video generators, the observed TVD
averages 31.2. Samples drawn from the reference distributions average 8.33 TVD,
and 99\% of the simulated averages remain below 9.22. Each generator's observed
TVD also exceeds the 99th percentile of its own matched baseline. Finite
sampling therefore contributes to measured TVD, but it is too small to explain
the observed calibration gap. Table~\ref{tab:finite-sample-null-sensitivity}
shows that the same conclusion holds under alternative aggregation and a
stricter minimum-readout requirement. This analysis concerns conditional
Calibration TVD and does not evaluate Coverage or outcome-readout failures.

\FloatBarrier
\begin{table*}[!t]
  \centering
  \caption{\textbf{Finite-sample null sensitivity.}
  We repeat the matched analysis with pooled passing cells and with a stricter
  $n_{\mathrm{readout}}\geq30$ threshold. The primary and stricter analyses
  weight models equally; the pooled analysis weights model--scene cells
  equally. All rows use $B=50{,}000$ replicates, and $p_{\mathrm{MC}}$ is the
  plus-one upper-tail probability. These checks do not alter benchmark scores
  or model rankings.}
  \label{tab:finite-sample-null-sensitivity}
  \scriptsize
  \setlength{\tabcolsep}{5.0pt}
  \renewcommand{\arraystretch}{1.15}
  \begin{tabular}{@{}lrrrrrrr@{}}
    \toprule
    \rowcolor{blue!8}
    \textbf{Aggregation} & \textbf{Models} & \textbf{Cells} & \textbf{Observed}
      & \textbf{Null mean} & \textbf{Null q99} & \textbf{Exceedances} & $\boldsymbol{p_{\mathrm{MC}}}$ \\
    \midrule
    Equal-model (primary) & 11 & 208 & 31.2 & 8.33 & 9.22 & 0/50,000 & $2.0\times10^{-5}$ \\
    Pooled passing cells & 11 & 208 & 31.6 & 8.13 & 8.90 & 0/50,000 & $2.0\times10^{-5}$ \\
    Equal-model, $n_{\mathrm{readout}}\geq30$ & 11 & 159 & 31.8 & 6.67 & 7.44 & 0/50,000 & $2.0\times10^{-5}$ \\
    \bottomrule
  \end{tabular}
\end{table*}

\FloatBarrier

\subsection{Full Causal and Non-Causal Controls}
\label{app:causal-cue-full}

\begin{figure}[!htbp]
  \centering
  \includegraphics[width=\linewidth]{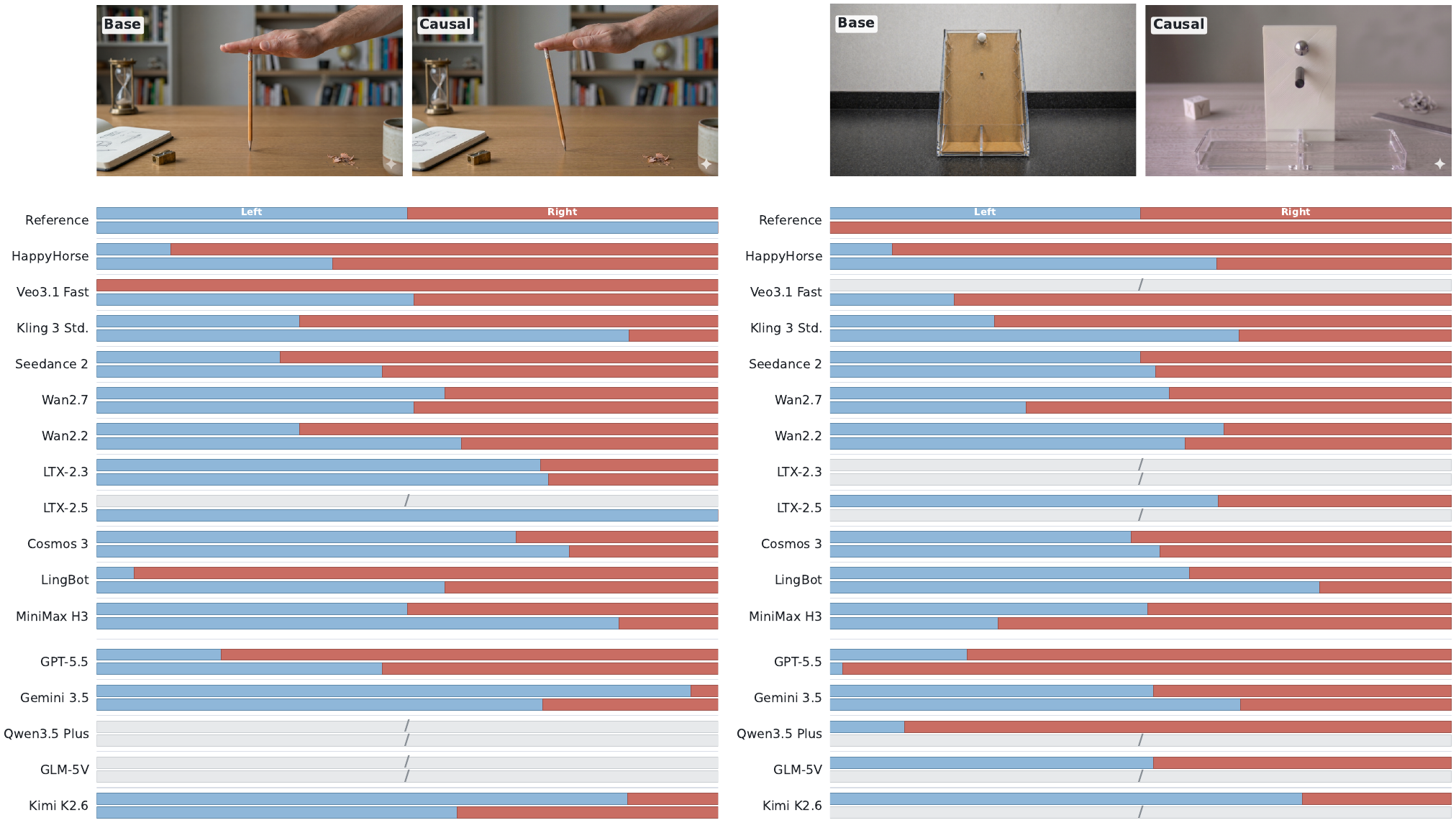}
  \caption{\textbf{Causal-state controls across video generators and direct VLM future samplers.} Each panel compares the outcome distribution for a base scene (upper bar) with that for a causal variant (lower bar), whose reference distribution changes. Bars report distributions conditional on readable, in-schema outcomes; gray bars marked / denote scenes that fail the outcome-readout gate.}
  \label{fig:causal-cue-full-causal}
\end{figure}

\begin{figure}[!htbp]
  \centering
  \includegraphics[width=\linewidth]{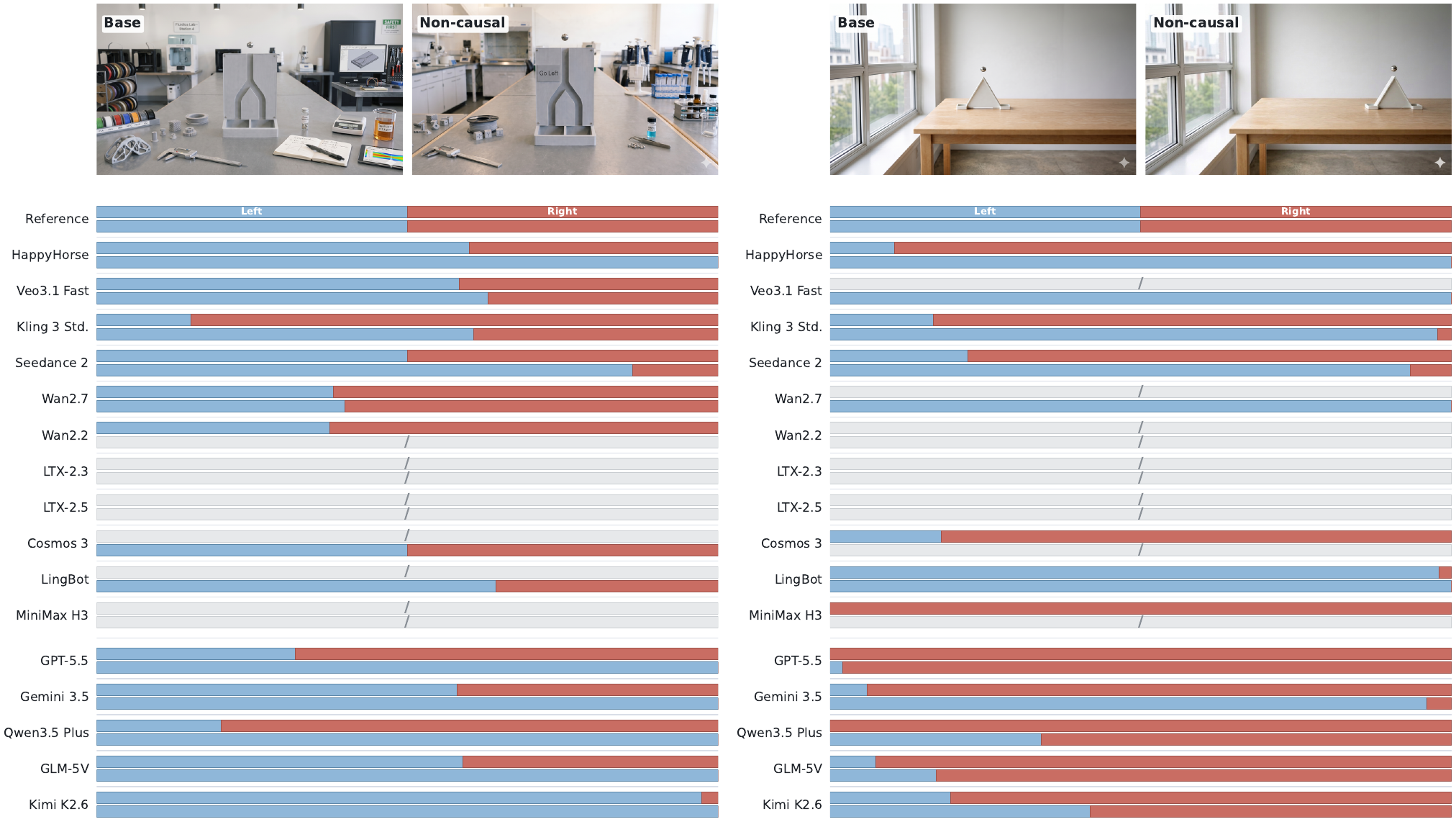}
  \caption{\textbf{Non-causal routing controls.} The paired scenes change an irrelevant cue while preserving the physical transition and reference distribution. Upper and lower bars show the base and cue-perturbed outcome distributions, respectively; gray bars marked / denote scenes that fail the outcome-readout gate.}
  \label{fig:causal-cue-full-routing}
\end{figure}

\begin{figure}[!htbp]
  \centering
  \includegraphics[width=\linewidth]{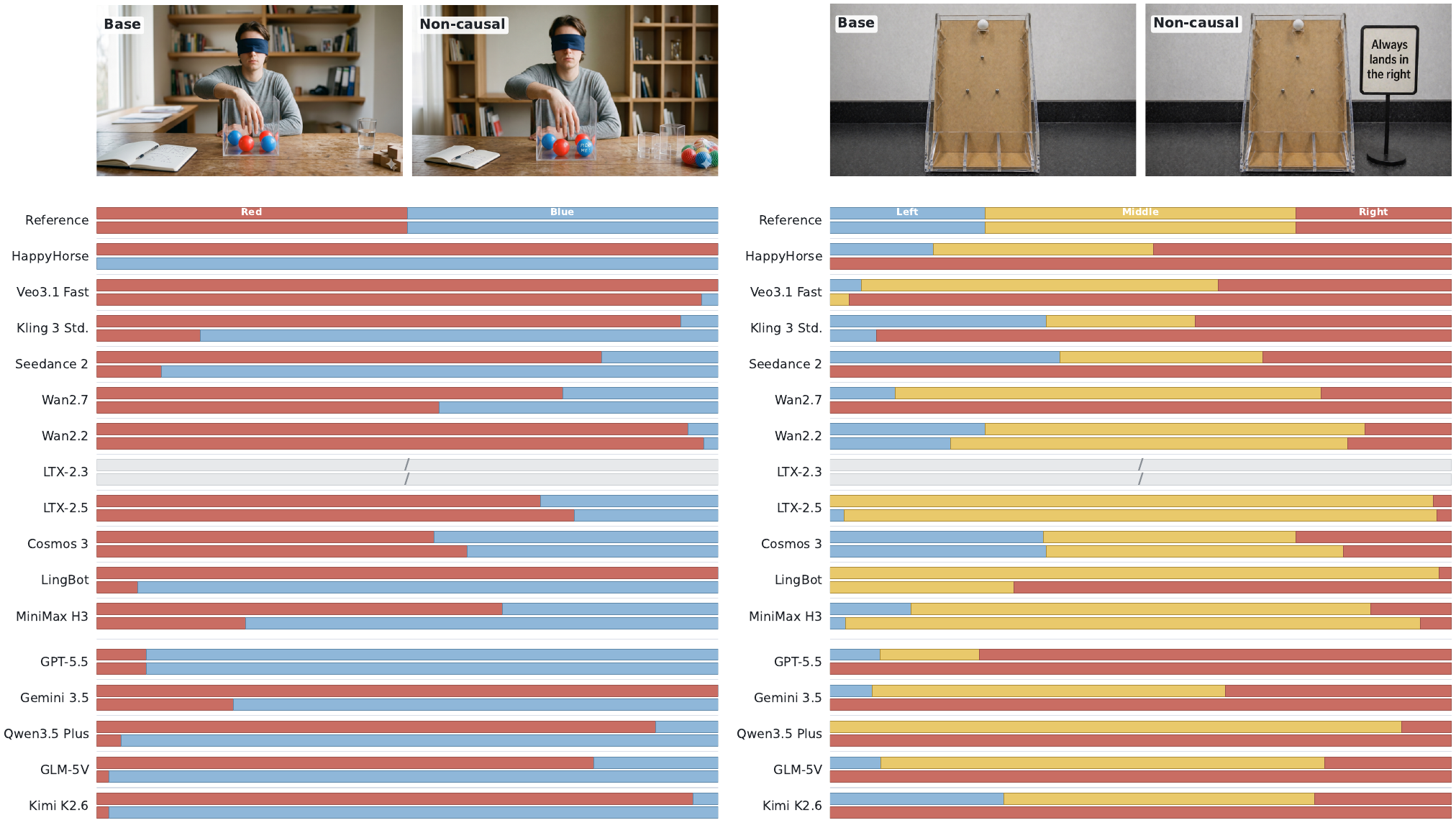}
  \caption{\textbf{Non-causal draw and text controls.} Distractor appearance in the blind-draw scene and an outcome-suggestive sign in the Galton-board scene leave the reference distribution unchanged. Several systems nevertheless move substantial probability mass between the paired conditions.}
  \label{fig:causal-cue-full-draw-text}
\end{figure}

\begin{figure}[!htbp]
  \centering
  \includegraphics[width=\linewidth]{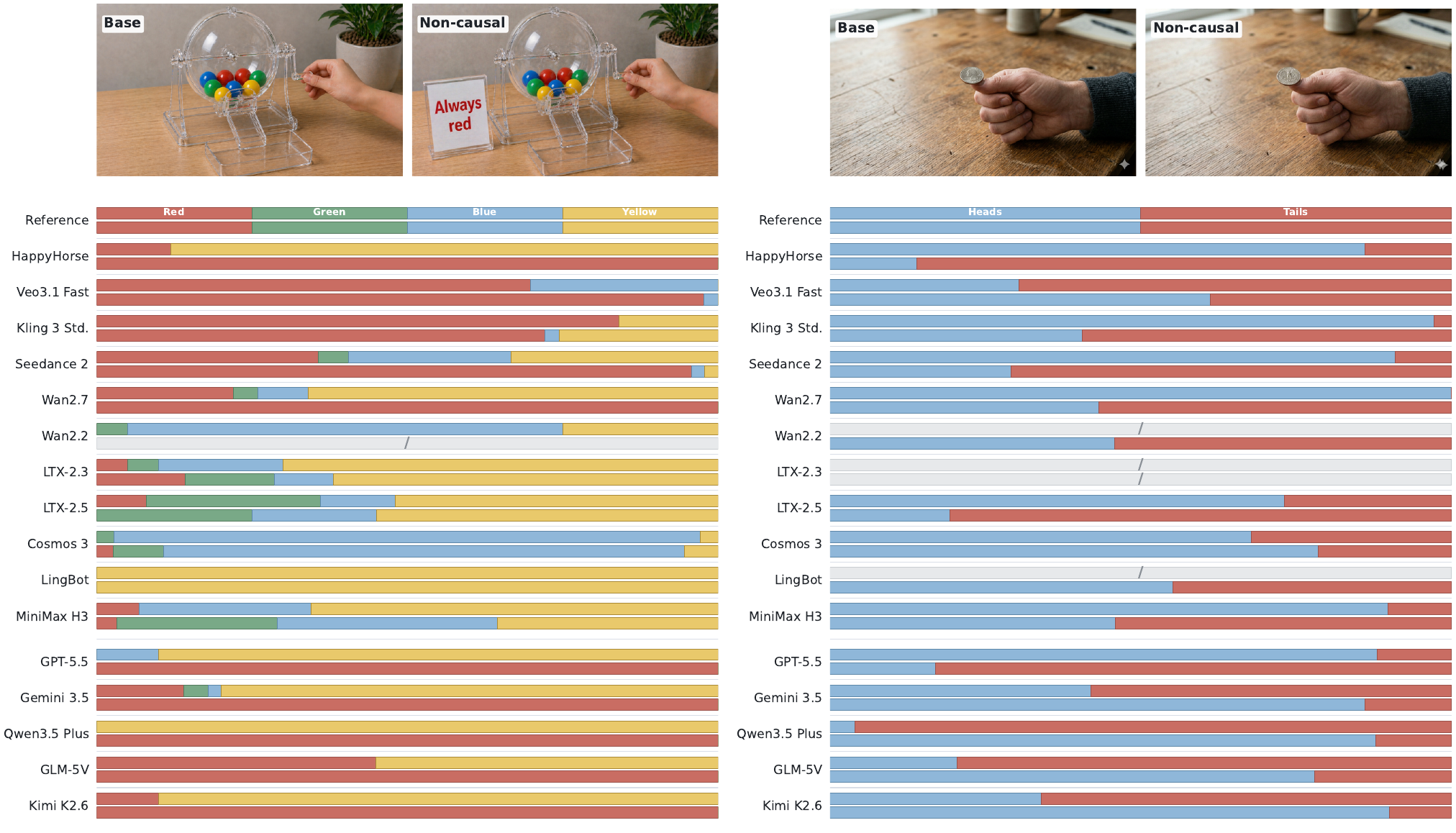}
  \caption{\textbf{Non-causal dispenser and coin controls.} Outcome-suggestive text or an irrelevant appearance change leaves the underlying chance process and reference distribution unchanged. Upper and lower bars show the base and cue-perturbed distributions conditional on readable, in-schema outcomes.}
  \label{fig:causal-cue-full-dispenser-coin}
\end{figure}

Figs.~\ref{fig:causal-cue-full-causal}--\ref{fig:causal-cue-full-dispenser-coin} expand the paired-control diagnostic in Fig.~\ref{fig:causal-cue-control} to eight controlled pairs. The roster contains all eleven video generators from the main evaluation and five vision-language models that sample future outcomes directly from the same initial observation and action. For every system and scene, we draw $K=50$ samples and project readable responses into the same fixed outcome set. Video-generator rows use PAWEval outcome readout on generated videos, whereas VLM rows use projected semantic predictions. The two groups therefore support a diagnostic comparison of paired response direction, not a cross-modality performance ranking.

Within each system, the upper bar shows the base scene and the lower bar shows its causal or non-causal variant. Bars report $P(\text{outcome}\mid\text{readable, in-schema})$ when the scene passes the same outcome-readout gate used in the main evaluation; gray bars marked / denote gate failures. A causal intervention changes the physical transition and reference distribution, so the model distribution should change accordingly. A non-causal intervention preserves both, so the distribution should remain stable.

\FloatBarrier

\section{Human Study and PAWEval Alignment}
\label{app:human-study}

The human study asks whether PAWEval and a human panel assign the same terminal
outcome to the same generated video. It audits PAWEval's outcome readout; it
does not validate PAWBench's TVD, Coverage, or model rankings.

\textbf{Study sampling.} The frozen study frame contains 1,500 generated videos from 29 PAWBench
scenes. Of these, 1,200 were selected to balance scene coverage, 150 to
strengthen within-scene model comparisons, and 150 to probe cases expected to
be difficult for the evaluator. Each video is one study item and receives
seven independent judgments from a screened Rapidata audience. The resulting
agreement therefore characterizes this frozen study frame rather than all
PAWBench rollouts.

\textbf{Annotation and comparison protocol.} Each Rapidata task\footnote{\url{https://www.rapidata.ai/}} shows one generated
video and asks annotators to identify its final visible outcome using a
scene-specific closed-choice question. The options express the canonical
PAWBench outcomes in human-readable terms and include \textit{Cannot tell} for
unclear, hidden, or unreadable endpoints. Annotators do not see PAWEval's
prediction or model metadata. We map their responses back to the canonical
outcome space and define the human label as decisive when the seven votes have
a unique modal outcome other than \textit{Cannot tell}; ties and
\textit{Cannot tell} modes remain non-comparable. A PAWEval label is comparable
when it is an in-schema physical outcome. Agreement is an exact canonical-label
match on the intersection of these two conditions.

The two eligibility conditions are parallel filters over the same 1,500
videos: humans provide a decisive label on 1,128 videos, and PAWEval provides
an in-schema label on 1,024. Their intersection contains 888 comparable videos.

\textbf{Agreement analysis.} Tab.~\ref{tab:human-study-agreement} reports 81.3\% exact agreement: PAWEval matches the decisive human label on 722 of the 888 comparable videos. The 888
videos are a subset of the full 1,500-video study frame, so this result is not
an all-row accuracy estimate.

Descriptively, agreement rises with the strength of the human mode: 58.7\% for
three or four matching votes, 77.3\% for five, and 91.7\% for six or seven.
This stratification shows where PAWEval--human agreement is concentrated; it
does not extend the 81.3\% result beyond the comparable subset. Overall, the
study supports a bounded conclusion: PAWEval often agrees with human judgment
when both produce clear terminal-outcome labels. PAWBench calibration and
coverage remain defined by the repeated-rollout metrics in
Appendix~\ref{app:metric-design}.

\begin{center}
\begin{minipage}{\linewidth}
\centering
\small
\captionof{table}{\textbf{PAWEval--human agreement on comparable videos.} Agreement is
the exact canonical-label match when both PAWEval and the seven-vote human
panel provide a clear physical outcome. Consensus rows are descriptive strata
of the 888-video comparison set.}
\label{tab:human-study-agreement}
\setlength{\tabcolsep}{6pt}
\begin{tabular}{lrrr}
\toprule
Human-consensus subset & Videos & Exact matches & Agreement \\
\midrule
Overall & 888 & 722 & 81.3\% \\
\midrule
Low consensus (3--4 votes) & 213 & 125 & 58.7\% \\
Moderate consensus (5 votes) & 154 & 119 & 77.3\% \\
High consensus (6--7 votes) & 521 & 478 & 91.7\% \\
\bottomrule
\end{tabular}
\end{minipage}
\end{center}

\section{Intervention Experiments}
\label{app:intervention-experiments}

Section~\ref{sec:alignment-interventions} intervenes at three interfaces where
future distributions may be shaped: the language supplied to the generator,
the initial noise used to sample it, and the training distribution absorbed by
the model. This appendix specifies how each intervention is constructed and
which quantities are held fixed. Quantitative comparisons are reported beside
their corresponding protocols.
% Native PE and SSoT were previously documented as language-side diagnostics.
% Their prose, tables, and figure are retained below as commented source but
% excluded from the current manuscript.

\begin{table}[!htbp]
  \centering
  \caption{\textbf{Experimental settings for the Section 5 probes.}
  Each row records the evaluated systems, prompt interface, sampling budget,
  and readout; the corresponding constructions follow below.}
  \label{tab:section5-experimental-settings}
  \scriptsize
  \setlength{\tabcolsep}{2.2pt}
  \renewcommand{\arraystretch}{1.15}
  \begin{tabular}{
    @{}
    >{\raggedright\arraybackslash}p{0.105\linewidth}
    >{\raggedright\arraybackslash}p{0.18\linewidth}
    >{\raggedright\arraybackslash}p{0.22\linewidth}
    >{\raggedright\arraybackslash}p{0.18\linewidth}
    >{\raggedright\arraybackslash}p{0.255\linewidth}
    @{}
  }
    \toprule
    \rowcolor{blue!8}
    \textbf{Probe} & \textbf{Systems} & \textbf{Prompt interface}
      & \textbf{Sampling} & \textbf{Readout and scope} \\
    \midrule
    VLM future sampling
      & Qwen3.5 Plus, GPT-5.5, GLM-5V Turbo, Kimi K2.6, Gemini 3.5 Flash
      & Fixed VLM scaffold; source image and action prompt are inserted at runtime
      & 25 scenes per track; $K=50$ queries; hosted-model randomness
      & Fixed GPT-5.5 projection; a scene fails above 30 unmapped or invalid responses \\
    \midrule
    PE / Oracle PE
      & Wan2.2, Cosmos 3 Super I2V, MiniMax H3, LTX-2.5
      & GPT-5.5 predicts an outcome and writes the PE prompt; Oracle PE manually places scheduled targets in the prompts
      & 25 scenes per track; $K=50$ rollouts under each condition
      & PAWEval target hit plus Calibration TVD or Coverage; all current model--track pairs \\
%     \midrule
%     Native PE
%       & Model-specific processors in Tab.~\ref{tab:model-specific-prompt-processing}
%       & Official prompt processor; no shared rewrite template
%       & $K=50$ rollouts per reported scene and condition
%       & PAWEval; only generator--condition pairs with completed results are reported \\
%     \midrule
%     SSoT
%       & Wan2.2, LTX-2.3
%       & Fixed controller selects one future and returns one final I2V prompt
%       & One controller call per rollout; $K=50$ rollouts per scene
%       & PAWEval; available generator--condition pairs only \\
    \midrule
    C2C
      & Wan2.2, LTX-2.3, Cosmos 3 Super I2V
      & Base prompt unchanged
      & $G=10$ groups of $m=5$; matched $K=50$ IID gallery
      & Common PAWEval protocol over all 25 scenes in both tracks \\
    \midrule
    LoRA
      & Wan2.2 I2V-A14B with five rank-32 adaptations
      & Direction-neutral pencil prompt
      & $K=50$ matched seeds for two physical states
      & PAWEval outcome counts; dose--response claims use adapted models, with Base as reference \\
    \bottomrule
  \end{tabular}
\end{table}

\FloatBarrier

\subsection{Language-Side Diagnostics and Interventions}
\label{app:language-intervention-diagnostics}

We evaluate language at two interfaces: predicting possible outcomes without
video synthesis and requesting a specific outcome from a video generator. In
PE, GPT-5.5 predicts a possible outcome from the initial observation and action
without access to the reference distribution, then writes a generator prompt
that requests this outcome. In Oracle PE, we instead write the target outcome
directly into the generator prompt. These settings separate two errors:
predicting the wrong outcome distribution and failing to produce a requested
outcome in video.

\FloatBarrier
\begin{figure}[!ht]
  \centering
  \includegraphics[width=\linewidth]{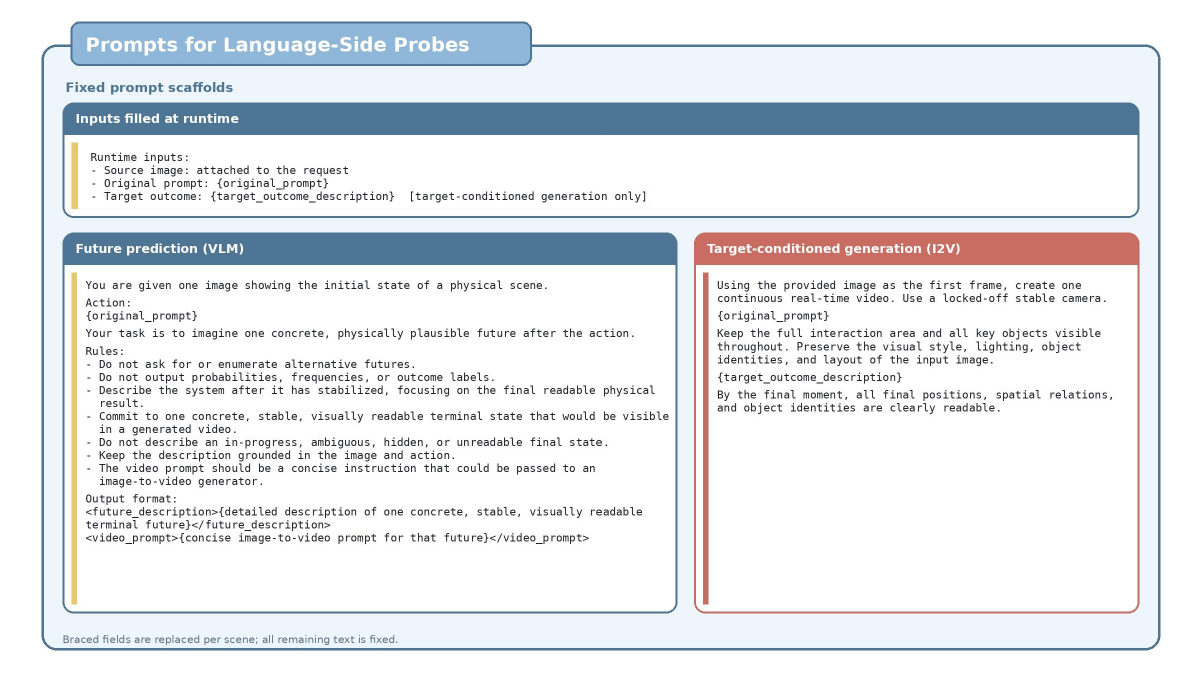}
  \caption{\textbf{Prompts for the two language-side probes.}
  The VLM scaffold requests one plausible future without exposing outcome
  labels or probabilities. The target-conditioned I2V scaffold instead adds
  one terminal outcome to the original scene prompt. Braced fields are filled
  at runtime.}
  \label{fig:language-prompt-interfaces}
\end{figure}

\FloatBarrier

\noindent\textbf{VLM sampling without video.}
We first test whether repeated VLM predictions produce the required outcome
distribution before video generation. For each of the 25 scenes in both
tracks, we show a VLM the same source image and action prompt and ask it to
describe one possible outcome, repeating the query $K=50$ times. A fixed
GPT-5.5 projector maps these free-form predictions to the scene-specific
PAWBench outcome set, yielding an empirical distribution that can be evaluated
against the scene's outcome specification.

Only responses mapped to valid outcomes contribute to this conditional
distribution. Unmapped or invalid responses instead count toward the scene
gate, which fails a scene when they exceed 30 of the 50 queries. Over passing
scenes, Calibration reports valid-only TVD and Coverage reports support
recovery; SPR separately gives the fraction of the 25 scenes that pass. We
reuse the frozen responses and projections for both tracks rather than
regenerating or reprojecting them.

\noindent\textbf{Prompt Engineering(PE).}
For each rollout, GPT-5.5 predicts one possible outcome from the initial
observation and action, then writes a generator prompt requesting that outcome.
It receives neither the reference distribution nor a target outcome from the
benchmark.

\noindent\textbf{Oracle PE.}
For Oracle PE, we manually specify one PAWBench target outcome in every
generator prompt. For PAW-Calibration, we convert the reference probabilities
into $K=50$ target
requests using largest-remainder allocation. For PAW-Coverage, we divide the
50 requests as evenly as possible across the valid outcomes. We deterministically
permute each target schedule across rollout indices and reuse the same schedule
for every generator. Across the 25 scenes in each track, this gives 2,500
target-conditioned rollouts for each generator.
PAWEval reads the terminal outcome, and a rollout counts as a target hit only
when that outcome matches the request; unreadable, invalid, missing, and
different outcomes are all misses. We then ask a separate distributional
question by comparing the resulting Calibration TVD and Coverage with matched
Base rollouts. Target-hit rate measures whether a generator follows an
individual outcome request, whereas the PAWBench metrics measure the aggregate
distribution produced across requests.

\iffalse
\noindent\textbf{Native PE.}
Native PE does not name a benchmark outcome. Instead, we invoke the prompt
processor distributed with each generator and pass its output through that
model's standard conditioning pathway
\citep{wan2025wan22,nvidia2025cosmos,agarwal2026cosmos,minimax2026h3,
lingbot2026video,hacohen2026ltx2,lightricks2026ltx25}. These processors produce different forms of
conditioning. Tab.~\ref{tab:model-specific-prompt-processing} summarizes these
interfaces, which include rewritten positive prompts, structured captions, and
paired positive--negative prompts.
Native PE is therefore a family of model-specific interfaces, not one shared
intervention whose effect can be pooled across generators.

\noindent\textbf{SSoT.}
SSoT uses a generated string as an internal source of randomness for a
language-model decision~\citep{misaki2025ssot}. We call the controller once per
rollout. Given the source image and action request, it generates and manipulates
a 24-character string, selects one plausible physical future, and rewrites the
request as a self-contained I2V prompt that is passed unchanged to the video
generator. Our controller receives neither a predefined outcome set nor a
target distribution. The string can vary which future is selected, but it
cannot reveal the reference probability mass.
\fi

% \input{figure/appendix/language_intervention_diagnostic}
% \FloatBarrier

Tab.~\ref{tab:target-prompt-results} reports Base, GPT-5.5 PE, and
Oracle PE results for Wan2.2, Cosmos 3 Super I2V, MiniMax H3, and LTX-2.5.
All conditions cover the 25 scenes in each track at $K=50$.

% \input{table/appendix/native_pe_across_generators}
% \FloatBarrier

\noindent\textbf{Language steers outcomes but does not determine their distribution.}
Tab.~\ref{tab:world-model-gap-diagnostics} shows that repeated VLM predictions
remain misaligned with the reference distribution.
Tab.~\ref{tab:target-prompt-results} shows that PE raises Calibration
TVD relative to Base for all four generators and improves Coverage for only
two, whereas Oracle PE lowers mean Calibration TVD and raises mean Coverage
for all four. Fig.~\ref{fig:explicit-target-realization}
nevertheless shows that only 37.6--58.1\% of their rollouts reach the requested
outcome. The results expose both errors: the language controller predicts the
wrong outcome distribution, and the video generator often fails to produce a
requested outcome. Language can change individual rollouts without reliably
aligning their aggregate distribution.

\subsection{Coupled Noise Sampling}
\label{app:c2c-experimental-details}

The initial noise of a diffusion generator is usually sampled independently
across rollouts. C2C asks a narrower question: can a finite gallery cover the
generator's accessible futures more evenly when these noises are coupled,
while preserving the standard-Gaussian marginal seen by every rollout?

\noindent\textbf{Marginal-preserving coupling.}
For group $g$, let $\epsilon_{g,1},\ldots,\epsilon_{g,m}$ be independent
standard-Gaussian noise tensors. Following the centered construction of
Couple to Control~\citep{jia2026coupletocontrol}, we form
\begin{equation}
  \epsilon_{g,i} \overset{\mathrm{i.i.d.}}{\sim} \mathcal{N}(0,I),
  \qquad
  \widetilde{\epsilon}_{g,i}
  = \sqrt{\frac{m}{m-1}}
  \left(
    \epsilon_{g,i}
    - \frac{1}{m}\sum_{j=1}^{m}\epsilon_{g,j}
  \right).
  \label{eq:c2c-group-construction}
\end{equation}
Each $\widetilde{\epsilon}_{g,i}$ remains marginally distributed as
$\mathcal{N}(0,I)$, while the samples within a group satisfy
$\sum_i\widetilde{\epsilon}_{g,i}=0$ and
$\operatorname{Cov}(\widetilde{\epsilon}_{g,i},
\widetilde{\epsilon}_{g,j})=-I/(m-1)$ for $i\neq j$. Groups are constructed
independently. The intervention therefore changes the joint distribution of
the gallery without changing the noise distribution of any single rollout.

\noindent\textbf{Experimental construction.}
We use $G=10$ independent groups with $m=5$ samples per group, yielding
$K=Gm=50$ rollouts for every model--scene pair. The coupled tensors are
constructed at each generator's native initial-latent shape and supplied at
the start of denoising. The matched IID condition instead draws $K$
independent tensors from the same $\mathcal{N}(0,I)$ marginal; the source
image, prompt, generator, and remaining generation settings are held fixed.
Tab.~\ref{tab:c2c-sampling-results} reports Wan2.2, LTX-2.3, and Cosmos 3 Super
I2V on the full 25-scene PAW-Calibration roster and the full 25-scene
PAW-Coverage roster.
All C2C and IID rollouts follow the common PAWEval outcome-readout and
aggregation protocol in Appendix~\ref{app:metric-design}; the intervention
changes only the dependence among initial noises, not the validity gate or
metric denominator.

Because C2C preserves the marginal distribution of every initial noise, it
does not alter the generator's learned one-rollout conditional law. It instead
probes whether a finite IID gallery misses futures that the generator can
already realize. Improved coverage under coupling is therefore evidence about
finite-budget exploration, not about learning a better-calibrated world model.

\subsection{Training-Distribution Intervention}
\label{app:training-intervention-details}

The previous interventions leave the generator unchanged. Here we instead ask
whether the relative frequency of two futures in the training data becomes
part of the generator's learned outcome distribution. We vary only the
left--right composition of the training dataset, while holding its size, the
underlying video pool, the training recipe, and the inference request fixed.

\noindent\textbf{Training mixtures.}
Starting from 100 left-falling and 100 right-falling videos, we form five
training datasets of 2,000 examples with left/right ratios of $0/100$,
$20/80$, $50/50$, $80/20$, and $100/0$. Source videos are repeated uniformly
within each direction, so the datasets differ only in the relative frequency
of the two outcomes. The three interior mixtures retain both futures and
isolate their relative frequency; the two endpoints serve as boundary
conditions. Every training example uses the same neutral motion prompt.

\noindent\textbf{LoRA adaptation.}
Each mixture adapts the same Wan2.2 I2V-A14B base model with rank-32
LoRA~\citep{hu2022lora}. Training uses $832\!\times\!480$ clips of 49 frames,
a learning rate of $10^{-4}$, and 2,000 optimizer updates per expert; all other
recipe choices are shared across mixtures. Because Wan2.2 divides denoising
between high- and low-noise experts, we train paired adapters for each mixture
and apply them to the corresponding experts at inference with unit LoRA
weight. This produces five adapted models; the original Wan2.2 model provides
an unadapted reference.

\noindent\textbf{Evaluation across physical states.}
We evaluate Base and the five LoRA conditions on two pencil scenes using the
same source image within each scene, a direction-neutral action prompt, and a
matched set of $K=50$ sampling seeds for every model--scene pair. The upright
pencil has a $50/50$ left--right reference,
whereas the initially left-leaning pencil has a $100/0$ reference. The five
adapted models share one generation profile. The Base samples use the same
scene, request, and seeds, but differ in output geometry and lack complete
diffusion-profile metadata. We therefore draw the dose--response comparison
from the LoRA models and use Base only to show the unadapted behavior. The two
scenes test whether the intervention writes a shared directional prior or a
distribution that changes with the observed physical state.

\begin{table}[!htbp]
  \centering
  \small
  \setlength{\tabcolsep}{7pt}
  \renewcommand{\arraystretch}{1.10}
  \caption{\textbf{Outcome counts under the training-distribution intervention.}
  Each cell reports $L/R/\mathrm{invalid}$ over $K=50$; Fig.~\ref{fig:training-mass-response}
  plots $L/(L+R)$ over readable outcomes. Base
  denotes the unadapted model; dose--response comparisons use the LoRA rows.}
  \label{tab:training-distribution-outcomes}
  \vspace{0.3em}
  \begin{tabular}{@{}lccc@{}}
    \toprule
    \rowcolor{blue!8}
    \textbf{Profile} & \textbf{Training $L/R$}
      & \textbf{Upright pencil} & \textbf{Left-leaning pencil} \\
    \midrule
    Base & -- & 16/33/1 & 27/19/4 \\
    \midrule
    LoRA & 0/100   & 0/36/14  & 1/35/14 \\
    LoRA & 20/80   & 12/25/13 & 17/17/16 \\
    LoRA & 50/50   & 25/9/16  & 30/7/13 \\
    LoRA & 80/20   & 48/1/1   & 50/0/0 \\
    LoRA & 100/0   & 45/0/5   & 48/0/2 \\
    \bottomrule
  \end{tabular}
\end{table}

\FloatBarrier

Across the three interior mixtures, conditional left-fall frequency increases
in both scenes. Training composition therefore shifts mass among supported
futures, but the experiment establishes neither exact ratio recovery nor
scene-conditioned probability learning: both scenes respond similarly despite
their different reference distributions.

\section{Qualitative Examples and Failure Cases}
\label{app:qualitative-cases}

This appendix complements the aggregate PAWBench results with 13 case cards
from distinct scenes. Each card shows a compact movie strip from one rollout,
together with its scene, model, generation condition, instruction, and visible
behavior. The examples illustrate how PAWEval reads terminal outcomes and why
some visually plausible videos fail to complete the intended physical process.
They provide qualitative evidence and do not enter any reported aggregate
metric.

The set includes successful physical trials and typical failure modes:
action-execution failures, object-continuity breaks, physically inconsistent
trajectories, apparatus instability, and questionable but readable clips. These
cards illustrate the aggregate patterns in Tab.~\ref{tab:main},
Fig.~\ref{fig:failure-anatomy}, Tab.~\ref{tab:world-model-gap-diagnostics}, and
Fig.~\ref{fig:causal-cue-control}.

\begingroup
\makeatletter
\setlength{\@fptop}{0pt}
\setlength{\@fpsep}{12pt}
\setlength{\@fpbot}{0pt plus 1fil}
\makeatother
% Auto-generated by figure/scripts/build_appendix_d_case_cards.py; do not edit by hand.

\begin{figure}[!htbp]
\centering
\includegraphics[width=\linewidth]{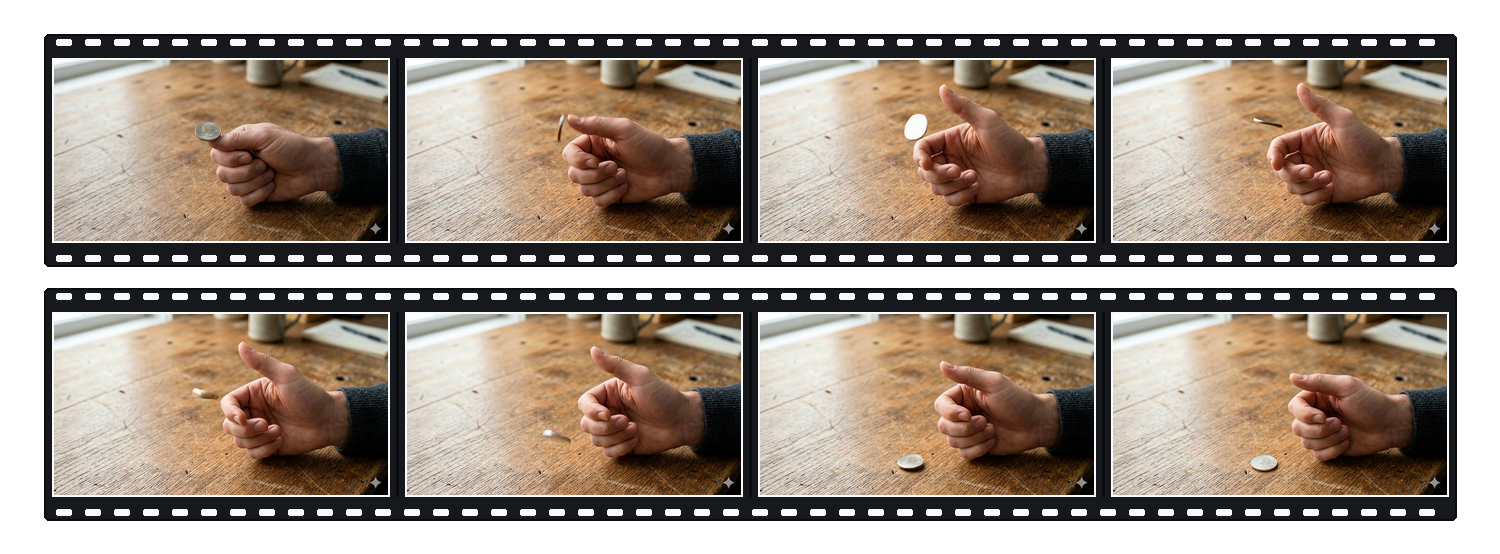}
\caption{\textbf{Example of the scene ``Coin flip'' generated by HappyHorse.} The model is instructed to flick the coin once. The generated rollout ends with the coin lying heads-up on the table.}
\label{fig:appendix-case-01}
\end{figure}

\begin{figure}[!htbp]
\centering
\includegraphics[width=\linewidth]{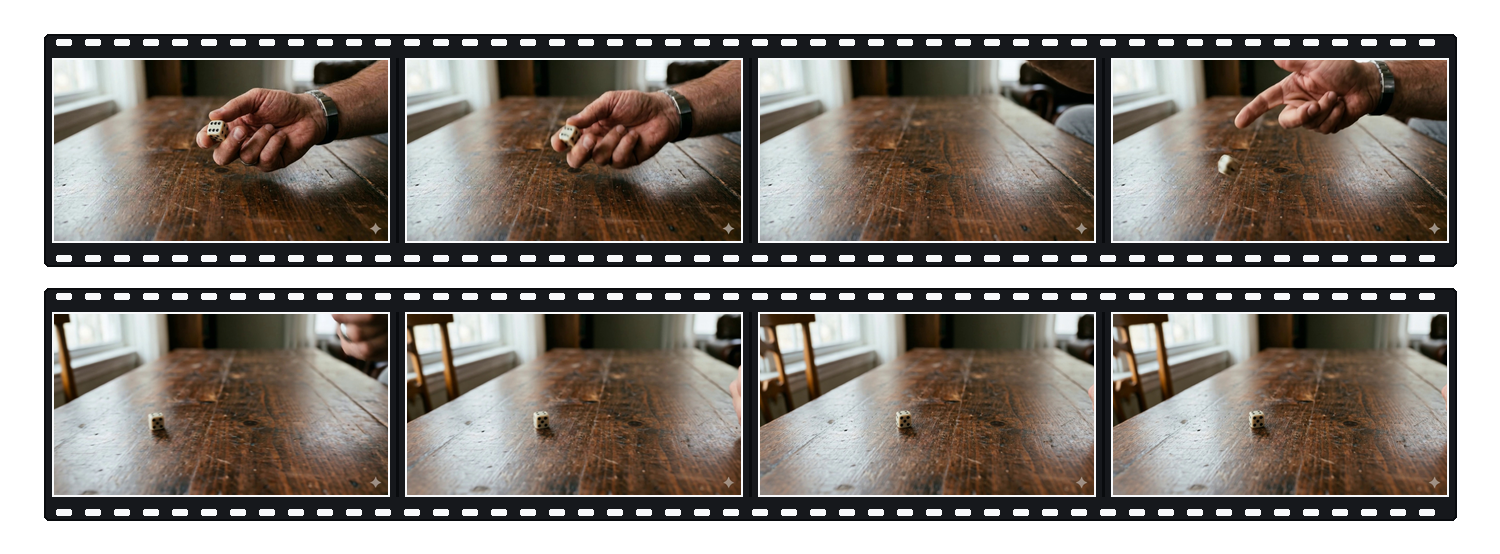}
\caption{\textbf{Example of the scene ``Die odd/even toss'' generated by Kling 3 Std.} The model is instructed to throw the die once onto the table. The generated rollout ends with the die settled on the table, showing five pips on top.}
\label{fig:appendix-case-02}
\end{figure}

\begin{figure}[!htbp]
\centering
\includegraphics[width=\linewidth]{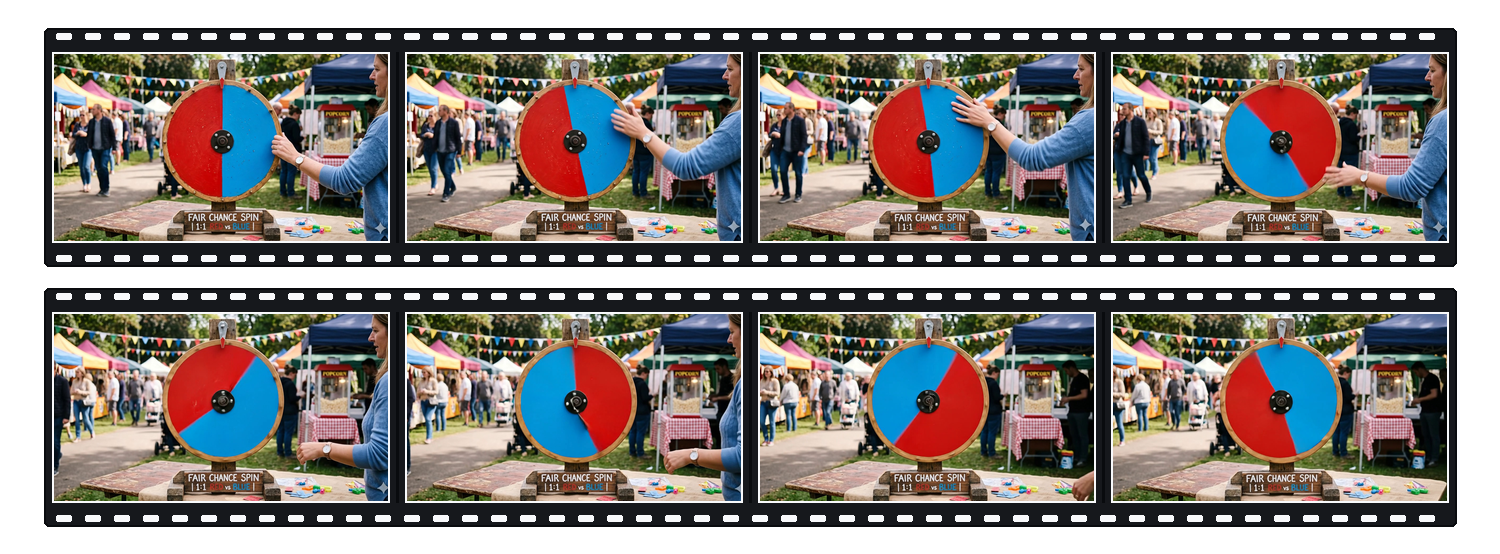}
\caption{\textbf{Example of the scene ``Two-color equal-sector spinner'' generated by Veo3.1 Fast.} The model is instructed to spin the wheel once under the fixed pointer. The final frame places the blue sector under the pointer, but the spin is visually inconsistent across the rollout.}
\label{fig:appendix-case-03}
\end{figure}

\begin{figure}[!htbp]
\centering
\includegraphics[width=\linewidth]{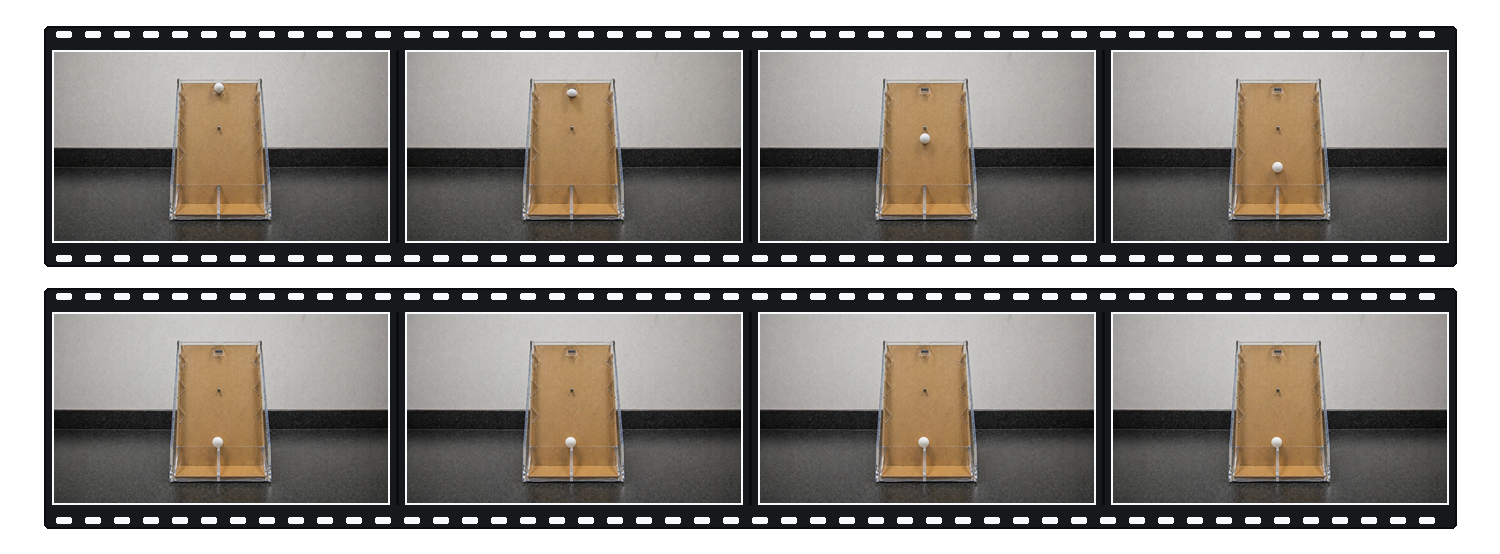}
\caption{\textbf{Example of the scene ``One-peg Galton board'' generated by Wan2.2.} The model is instructed to release the ball once. The ball settles on the divider instead of entering either bin.}
\label{fig:appendix-case-04}
\end{figure}

\begin{figure}[!htbp]
\centering
\includegraphics[width=\linewidth]{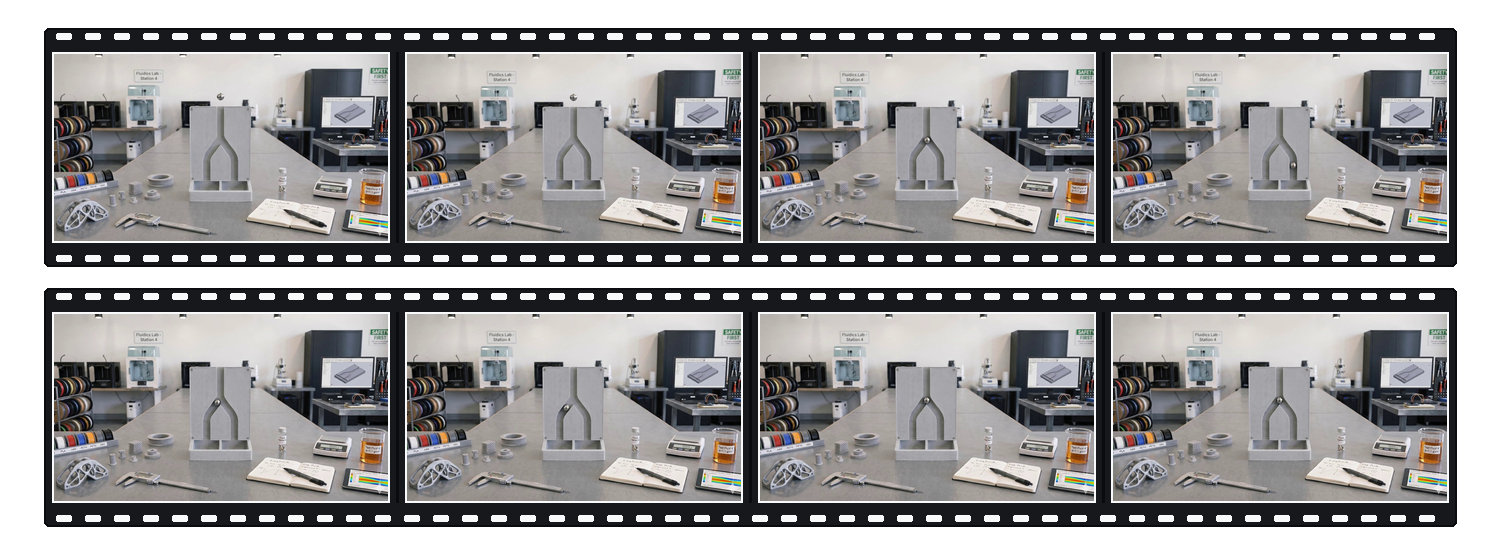}
\caption{\textbf{Example of the scene ``Y-track branch'' generated by Seedance 2.} The model is instructed to release the ball once. The ball remains stuck at the split rather than traveling down one branch.}
\label{fig:appendix-case-05}
\end{figure}

\begin{figure}[!htbp]
\centering
\includegraphics[width=\linewidth]{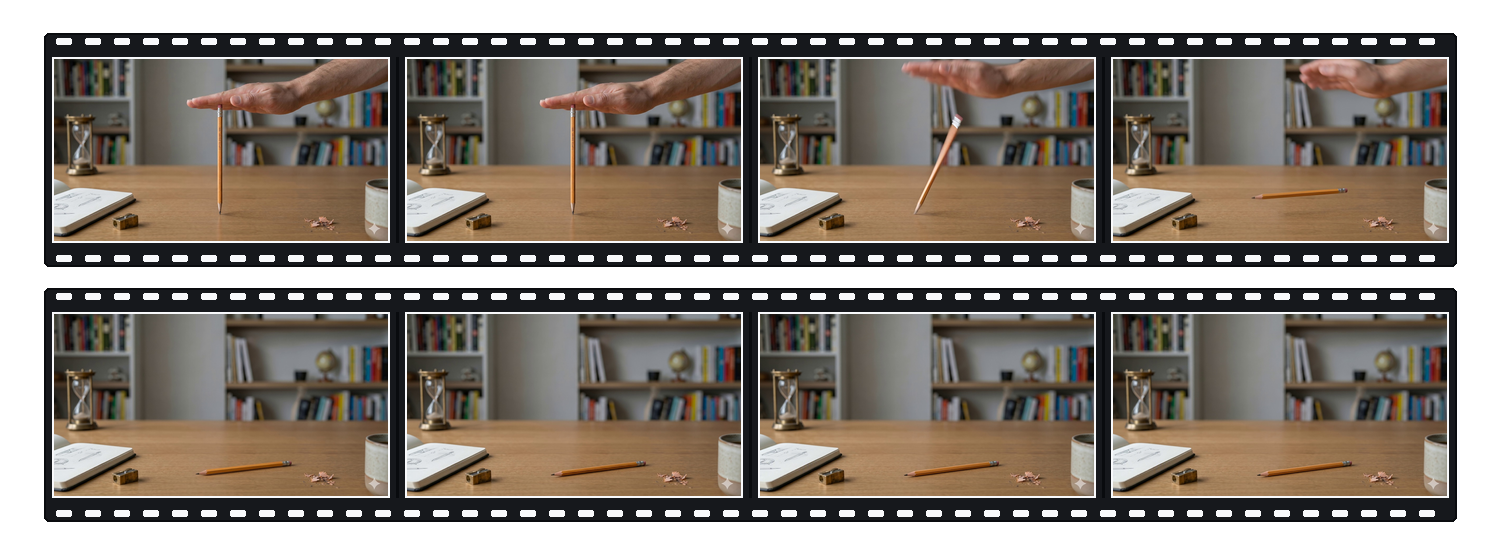}
\caption{\textbf{Example of the scene ``Vertical pencil fall'' generated by Wan2.2.} The model is instructed to move the hand upward once and let the pencil fall. The generated rollout ends with the pencil falling to the right.}
\label{fig:appendix-case-06}
\end{figure}

\begin{figure}[!htbp]
\centering
\includegraphics[width=\linewidth]{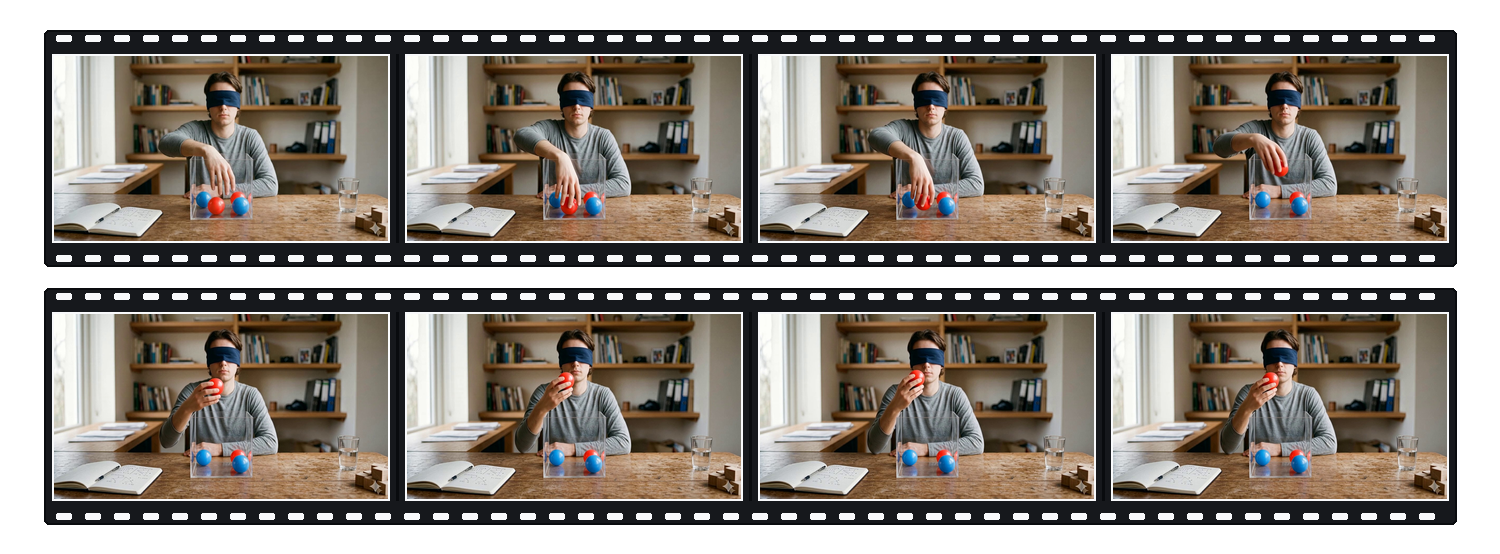}
\caption{\textbf{Example of the scene ``Blind ball draw'' generated by Wan2.7.} The model is instructed to pick up exactly one ball once. The person selects a red ball and holds it up by the end of the rollout.}
\label{fig:appendix-case-07}
\end{figure}

\begin{figure}[!htbp]
\centering
\includegraphics[width=\linewidth]{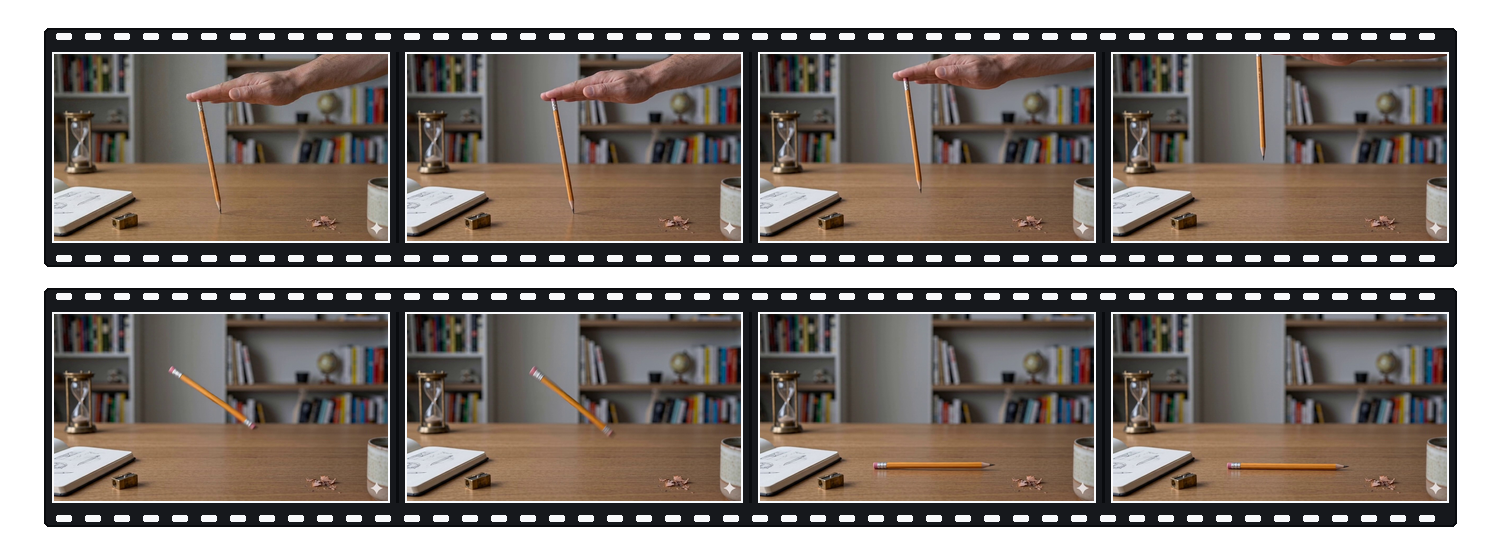}
\caption{\textbf{Example of the scene ``Left-Leaning Pencil Fall'' generated by Seedance 2.} The model is instructed to move the hand upward once and let the pencil fall. The pencil appears to fall left, but its identity is not preserved cleanly through the motion.}
\label{fig:appendix-case-08}
\end{figure}

\begin{figure}[!htbp]
\centering
\includegraphics[width=\linewidth]{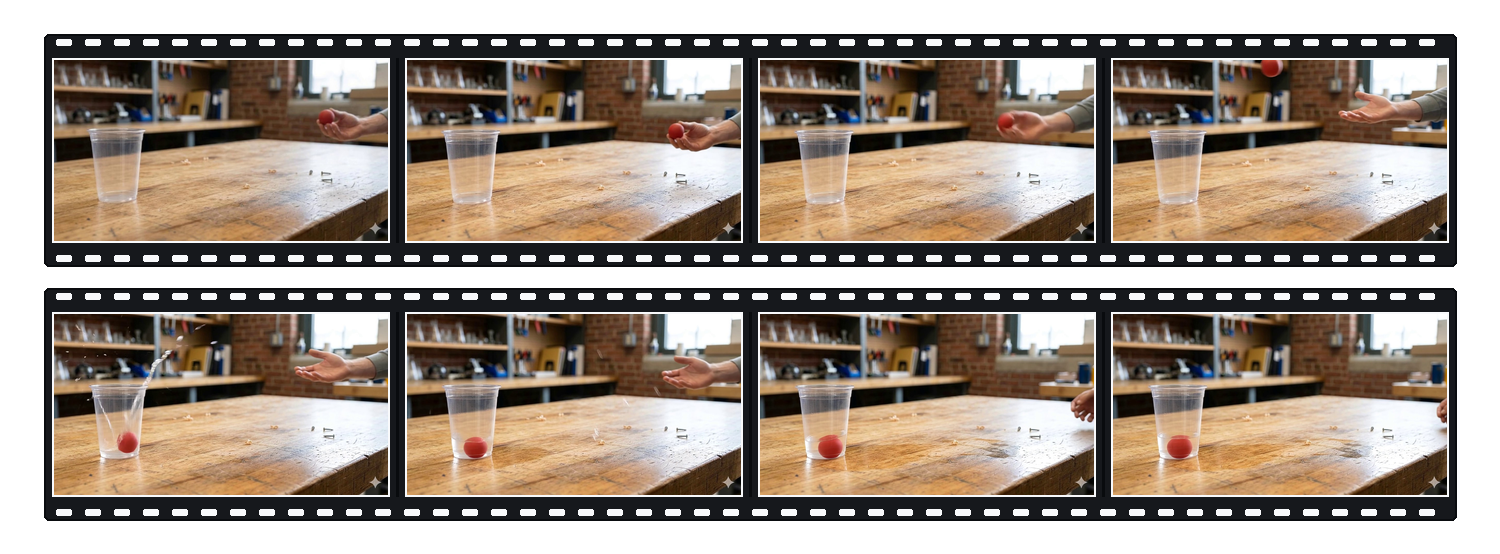}
\caption{\textbf{Example of the scene ``Ball Toss Into Cup'' generated by HappyHorse.} The model is instructed to toss the ball once from the visible hand pose toward the cup. The generated rollout ends with the ball landing cleanly inside the cup.}
\label{fig:appendix-case-09}
\end{figure}

\begin{figure}[!htbp]
\centering
\includegraphics[width=\linewidth]{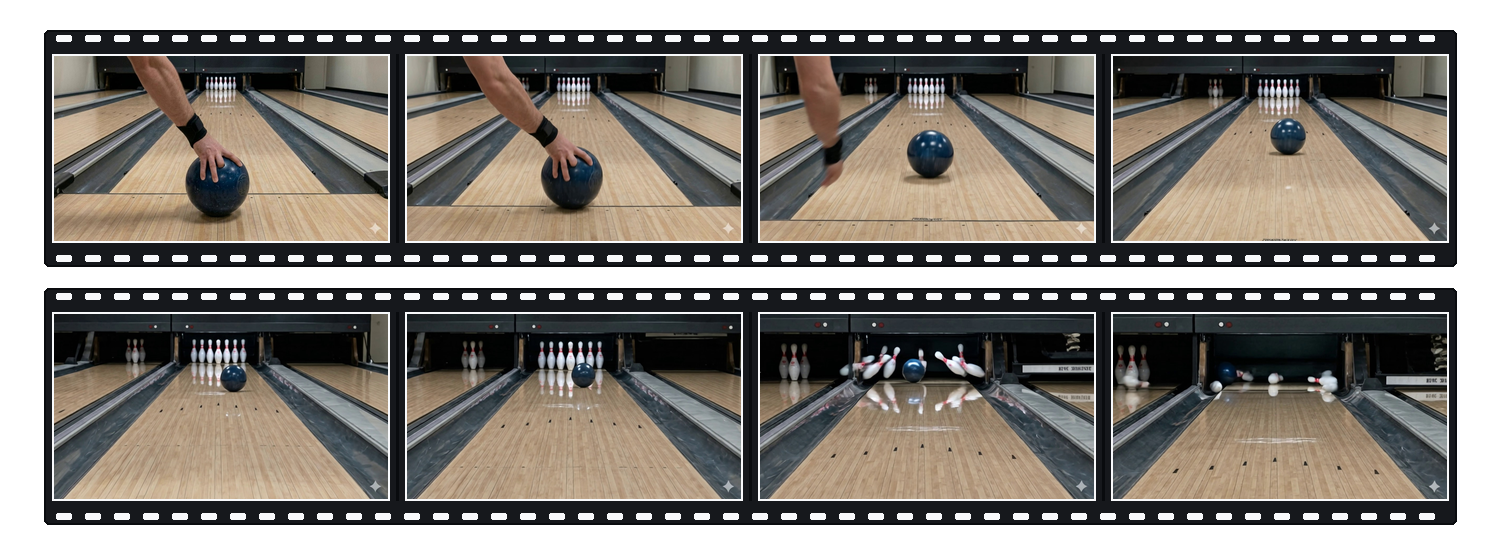}
\caption{\textbf{Example of the scene ``Seven-Pin Bowling Roll'' generated by Kling 3 Std.} The model is instructed to roll the bowling ball once down the lane. The generated rollout ends with all seven pins knocked down.}
\label{fig:appendix-case-10}
\end{figure}

\begin{figure}[!htbp]
\centering
\includegraphics[width=\linewidth]{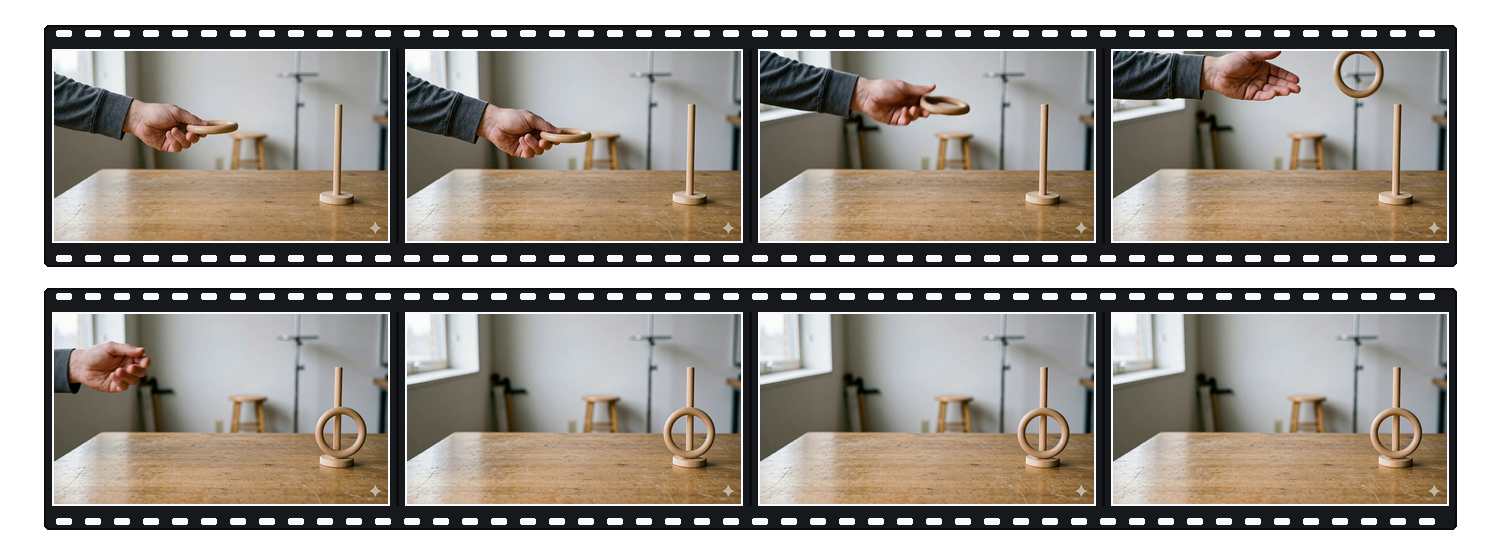}
\caption{\textbf{Example of the scene ``Ring toss toward peg'' generated by MiniMax H3.} The model is instructed to toss the ring once from the visible hand pose toward the peg. The ring travels toward the peg and remains around it in the final frames.}
\label{fig:appendix-case-11}
\end{figure}

\begin{figure}[!htbp]
\centering
\includegraphics[width=\linewidth]{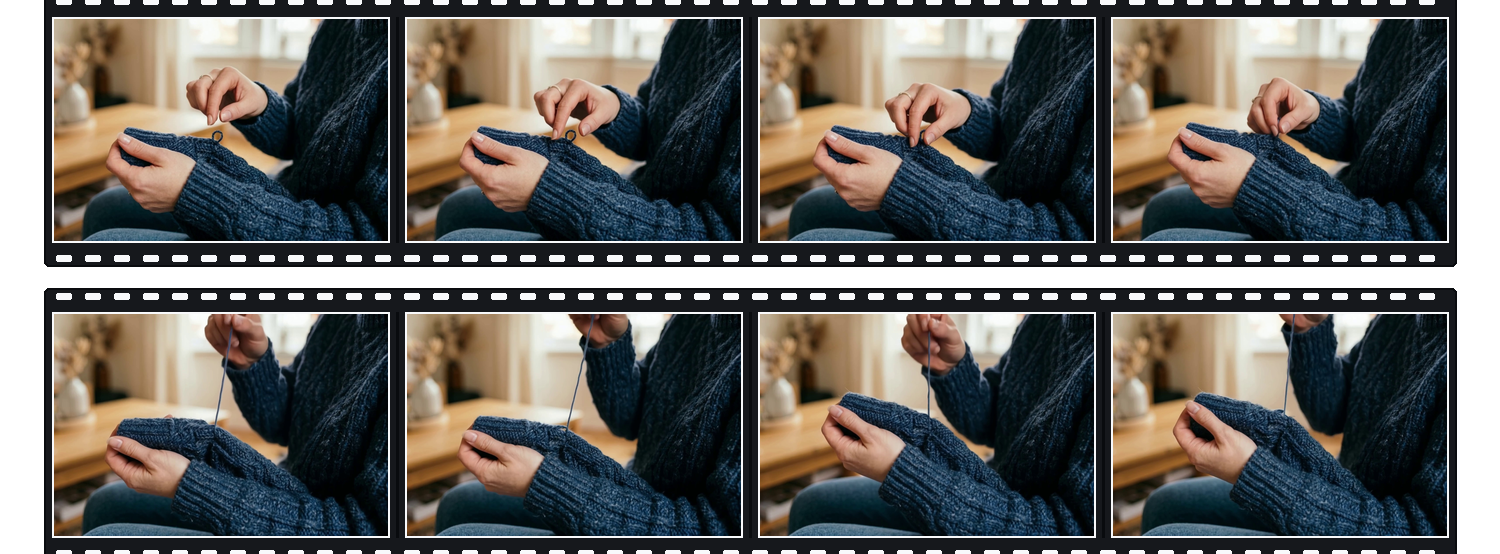}
\caption{\textbf{Example of the scene ``Loose-yarn pull'' generated by LTX-2.5 Base.} The model is instructed to pull the loose yarn end once. The loose yarn is pulled upward and visibly extends from the knitted fabric.}
\label{fig:appendix-case-12}
\end{figure}

\begin{figure}[!htbp]
\centering
\includegraphics[width=\linewidth]{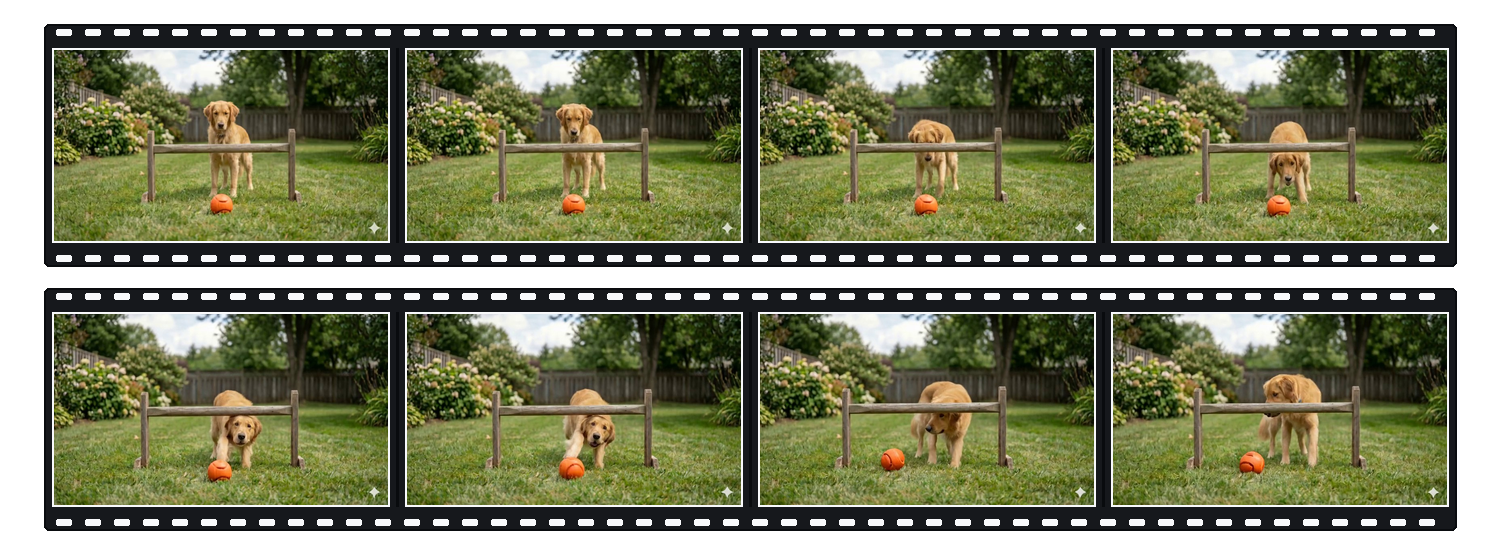}
\caption{\textbf{Example of the scene ``Dog Route Choice Toward Toy'' generated by Cosmos 3 Super I2V.} The model is instructed to let the dog make one reach attempt toward the toy. The dog lowers its head under the bar, then backs away and remains behind the obstacle.}
\label{fig:appendix-case-13}
\end{figure}

\FloatBarrier
\endgroup

\end{document}